\documentclass[letterpaper,10pt,journal,twoside]{IEEEtran}

\usepackage{amsmath,amsfonts}
\usepackage{algorithmic}
\usepackage{array}
\usepackage{textcomp}
\usepackage{stfloats}
\usepackage{url}
\usepackage{verbatim}
\usepackage{graphicx}
\def\BibTeX{{\rm B\kern-.05em{\sc i\kern-.025em b}\kern-.08em
    T\kern-.1667em\lower.7ex\hbox{E}\kern-.125emX}}
\usepackage{balance}
\usepackage{graphics}
\usepackage{subfigure}
\usepackage{booktabs}
\usepackage{multirow}
\usepackage{cite}
\usepackage{tikz,xcolor,hyperref}
\usepackage{algorithm}
\usepackage{fontawesome}
\definecolor{lime}{HTML}{A6CE39}
\DeclareRobustCommand{\orcidicon}{
	\begin{tikzpicture}
	\draw[lime, fill=lime] (0,0) 
	circle [radius=0.16] 
	node[white] {{\fontfamily{qag}\selectfont \tiny ID}};
	\draw[white, fill=white] (-0.0625,0.095) 
	circle [radius=0.007];
	\end{tikzpicture}
	\hspace{-2mm}
}

\foreach \x in {A, ..., Z}{
	\expandafter\xdef\csname orcid\x\endcsname{\noexpand\href{https://orcid.org/\csname orcidauthor\x\endcsname}{\noexpand\orcidicon}}
}

\begin{document}
\title{EV-MGRFlowNet: Motion-Guided Recurrent Network for Unsupervised Event-based Optical Flow with Hybrid Motion-Compensation Loss}

\author{Hao Zhuang\orcidA{}, Xinjie Huang\orcidB{}, Kuanxu Hou\orcidC{}, Delei Kong\orcidD{}, Chenming Hu\orcidE{}, and Zheng Fang\orcidF{}, \emph{Member, IEEE}
\thanks{
Manuscript received: April 1, 2023; Revised: April 1, 2023; Accepted: April 1, 2023. This work was supported by National Natural Science Foundation of China (62073066, U20A20197), the Fundamental Research Funds for the Central Universities (N2226001), and Intel Neuromorphic Research Community (INRC) Grant Award (RV2.137.Fang). \emph{(Corresponding author: Zheng Fang.)}

Hao Zhuang and Delei Kong are with College of Information Science and Engineering, Northeastern University, Shenyang 110169, China (e-mail: 2100922@stu.neu.edu.cn, kong.delei.neu@gmail.com).

Xinjie Huang, Kuanxu Hou, Chenming Hu and Zheng Fang are with Faculty of Robot Science and Engineering, Northeastern University, Shenyang 110169, China (e-mail: 2101979@stu.neu.edu.cn, 2001995@stu.neu.edu.cn, 2202039@stu.neu.edu.cn, fangzheng@mail.neu.edu.cn).

Digital Object Identifier (DOI): see top of this page.
}}
\maketitle

\begin{abstract}
Event cameras offer promising properties, such as high temporal resolution and high dynamic range. These benefits have been utilized into many machine vision tasks, especially optical flow estimation. Currently, most existing event-based  works use deep learning to estimate optical flow. However, their networks have not fully exploited prior hidden states and motion flows. Additionally, their supervision strategy has not fully leveraged the geometric constraints of event data to unlock the potential of networks. In this paper, we propose EV-MGRFlowNet, an unsupervised event-based optical flow estimation pipeline with motion-guided recurrent networks using a hybrid motion-compensation loss. First, we propose a feature-enhanced recurrent encoder network (FERE-Net) which fully utilizes prior hidden states to obtain multi-level motion features. Then, we propose a flow-guided decoder network (FGD-Net) to integrate prior motion flows. Finally, we design a hybrid motion-compensation loss (HMC-Loss) to strengthen geometric constraints for the more accurate alignment of events. Experimental results show that our method outperforms the current state-of-the-art (SOTA) method on the MVSEC dataset, with an average reduction of approximately 22.71\% in average endpoint error (AEE). To our knowledge, our method ranks first among unsupervised learning-based methods.
\end{abstract}

\section{Introduction}
\label{sec:introduction}

\begin{figure}[htbp]
\centering
\includegraphics[width=\columnwidth]{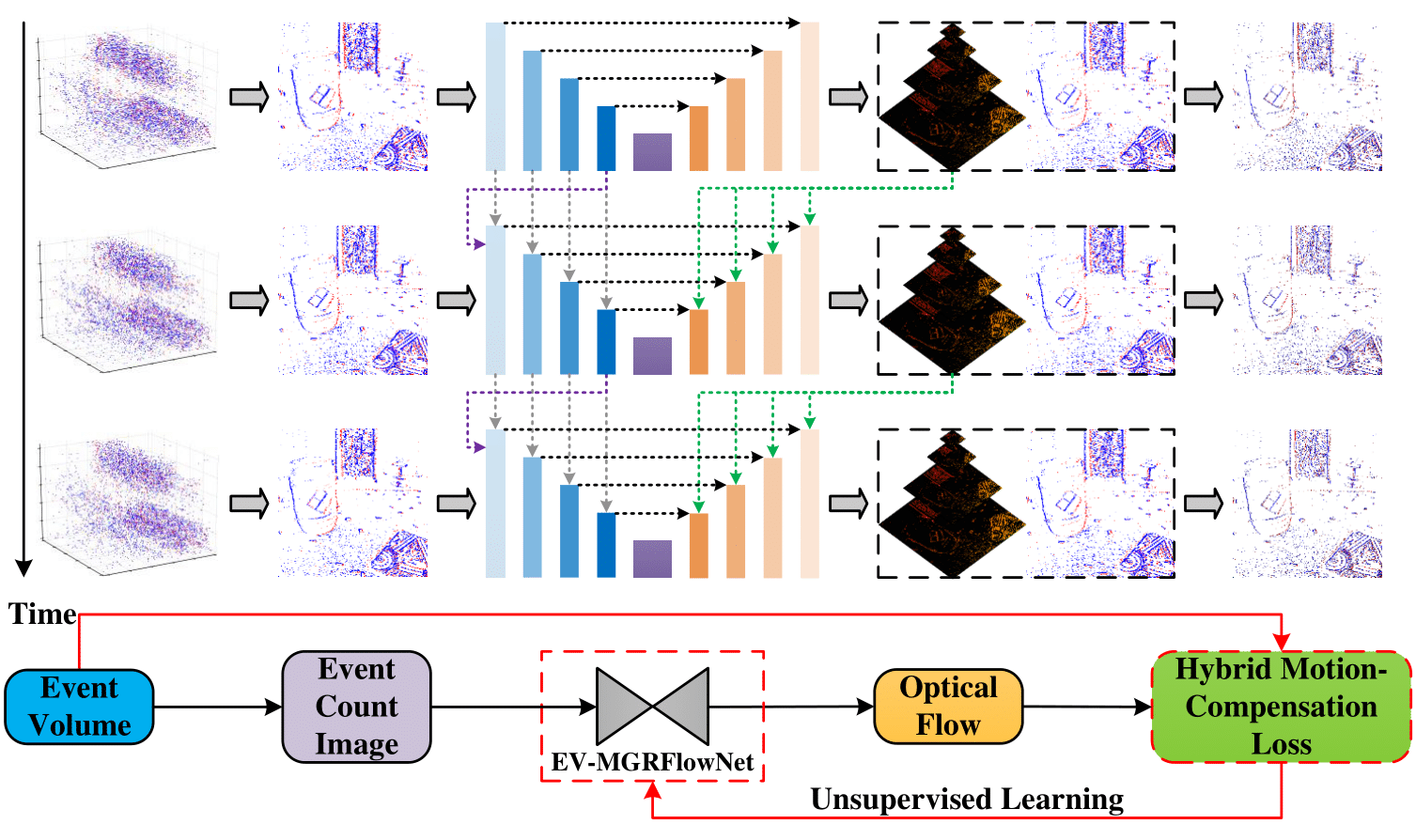}
\vspace{-2em}
\caption{Overview of Our EV-MGRFlowNet Pipeline. Our network can fully utilize prior crucial motion information which has been neglected by previous works to better capture motion patterns. Moreover, the network is trained by our proposed hybrid motion-compensation loss to allow for a more accurate alignment of events. See Fig. \ref{fig:2} for a more detailed pipeline.}
\label{fig:1}
\vspace{-1em}
\end{figure}

\IEEEPARstart{E}{vent} cameras are novel bio-inspired vision sensors that produce a series of events which are triggered by changes in log intensity with microsecond accuracy, thus having promising advantages of high temporal resolution and high dynamic range over frame-based cameras \cite{gallego2020event, chen2020event, zheng2023deep, kong2022event, hou2023fe, wu2021novel}. Nowadays, how to utilize these advantages into optical flow estimation has been a hot research topic. Optical flow estimation refers to estimating the velocity of objects on the image plane without scene geometry or motion \cite{baker2004lucas}. As a fundamental machine vision task, it has been widely used in areas such as image reconstruction\cite{paredes2021back}, visual odometry\cite{ban2020monocular}, object tracking\cite{liu2023motion} and autonomous driving\cite{liu2021automatic}.

Recently, there have been many works applying deep learning to event-based optical flow estimation and achieving excellent results \cite{zhu2018ev, zhu2019unsupervised, lee2020spike, paredes2019unsupervised, hagenaars2021self, gehrig2021raft, ding2022spatio, tian2022event, lee2022fusion}. And the aforementioned works have attempted to deal with two main challenges: (1) how to construct a network that can extract spatio-temporal association between neighboring events effectively to better capture motion patterns, and (2) how to design a loss function to unlock the network's potential through unsupervised learning of motion information. However, previous works have not reached consensus on how to fully resolve the above challenges.

To further resolve these challenges, we propose a method (EV-MGRFlowNet) \footnote{Supplementary Material: An accompanying video for this work is available at \url{https://youtu.be/RWlHCMIj6Ks}.} for unsupervised event-based optical flow estimation with motion-guided networks using a novel hybrid motion-compensation loss, and the whole pipeline is shown in Fig. \ref{fig:1}. First, we propose a novel recurrent encoder-decoder network that incorporates crucial information from the past, including hidden states and estimated flows. Second, we design a novel hybrid motion-compensation loss that allows for more accurate alignment of events. Finally, we evaluate our method on the Multi Vehicle Stereo Event Camera (MVSEC) \cite{zhu2018multivehicle} dataset and our method ranks first among the prior state-of-the-art (SOTA) unsupervised learning-based methods. Overall, the main contributions of this paper are as follows:
\begin{itemize}
\item We propose a novel event-based motion-guided recurrent network (EV-MGRFlowNet), which consists of a feature-enhanced recurrent encoder network (FERE-Net) and a flow-guided decoder network (FGD-Net), that can fully utilize prior hidden states and flows to capture motion patterns.
\item To achieve better event alignment by optical flow, we propose a novel hybrid motion-compensation loss (HMC-Loss) to strengthen the constraints of unsupervised learning, thus effectively improving the accuracy of optical flow estimation.
\item We comprehensively compare our EV-MGRFlowNet with other SOTA methods on the MVSEC dataset to demonstrate the advanced performance of our method, with an average reduction of approximately 22.71\% in average endpoint error (AEE).
\item The results of our ablation studies suggest that: prior hidden states and flows provide significant motion cues, and our proposed EV-MGRFlowNet can effectively utilize these information. Moreover, our proposed HMC-Loss enables effective unsupervised learning in multiple networks with various datasets.  
\end{itemize}

The rest of this article is organized as follows. Section \ref{sec:relatedwork}
reviews related work on event-based optical flow estimation. Section \ref{sec:methodology} describes the overall network pipeline of our EV-MGRFlowNet and introduces event representation, feature-enhanced recurrent encoder network, flow-guided decoder network and hybrid motion-compensation loss in detail. Next, the experimental results of EV-MGRFlowNet on the MVSEC dataset are presented in Section \ref{sec:experiments}. Finally, Section \ref{sec:conclusions} concludes this article.
\section{Related Work}
\label{sec:relatedwork}
Nowadays, deep learning is widely used for event-based optical flow estimation. Here, we briefly describe the existing works from the aspects of network architecture and training paradigm.

\textbf{Regarding network architecture}, early works \cite{zhu2018ev, zhu2019unsupervised} were mostly inspired by the frame-based optical flow pipeline, where each set of sparse events is transformed into a suitable image-like event representation and then fed into standard convolutional networks. The first work (EV-FlowNet) proposed by Zhu \cite{zhu2018ev} converted event volumes into 4-channel event images, extracted motion features through an encoder-decoder network and predicted multi-scale optical flow. Then, they extended it to the first unsupervised framework (EV-FlowNet+ \cite{zhu2019unsupervised}) that estimated optical flow, depth and ego-motion simultaneously. Although these works preserved spatio-temporal information of events by hand-crafted representations, using only convolutional networks can pose challenges in modeling spatio-temporal association between events. Some works introduced Spiking Neural Network (SNN) \cite{paredes2019unsupervised, lee2020spike, hagenaars2021self, lee2022fusion} or Convolutional Gated Recurrent Unit (ConvGRU) \cite{hagenaars2021self, ding2022spatio, tian2022event} to address this issue. The former works believe that the sequential dynamics of SNN can be used to naturally process events one-by-one, which helps to extract spatio-temporal association between events. Lee (Spike-FlowNet \cite{lee2020spike})  introduced a hybrid architecture that included a spiking encoder and a deep decoder to combine the benefits of both SNN and ANN, while Hagenaars et al. \cite{hagenaars2021self} integrated various spiking neuron models into the U-Net architecture directly. Following Spike-FlowNet, Lee \cite{lee2022fusion} further proposed Fusion-FlowNet  that processed events and frames simultaneously to leverage their complementarity. However, limited by the immaturity of gradient-based SNN training algorithms, current SNN-based methods are hard to surpass ANN-based methods \cite{ding2022spatio}. The latter works consider that ConvGRU enhanced the temporal modeling capabilities of the U-Net architecture with recurrent connections. Hagenaars \cite{hagenaars2021self} showed that a recurrent version of EV-FlowNet (ConvGRU-EV-FlowNet) could accumulate temporal information from previous slices of events. Then, Tian \cite{tian2022event} presented ET-FlowNet, a hybrid RNN-ViT architecture that used the transformer\cite{vaswani2017attention} to better learn the global context. Ding \cite{ding2022spatio} proposed a novel spatio-temporal recurrent network (STE-FlowNet) that allowed for a residual refine scheme\cite{hur2019iterative} with correlation layers. However, we find that existing ConvGRU-based networks have not adequately dug into spatio-temporal association that is essential for capturing motion patterns.

\textbf{Regarding training paradigm}, previous research can be broadly classified into two main categories: supervised and unsupervised learning methods. Supervised learning methods usually utilize the ground truth of optical flow for supervision. A typical work is E-RAFT proposed by Gehrig \cite{gehrig2021raft}, which introduced event cost volumes and achieved dense optical flow estimation through iterative refinement. However, considering supervised learning suffers from expensive annotation and inaccurate ground truth \cite{gehrig2021raft, shiba2022secrets}, unsupervised learning has attracted increasing interest, which fully exploits geometric constraints of event data while avoiding the aforementioned drawbacks. One type of method is based on photometric consistency loss between two consecutive intensity images with the brightness constancy assumption, which includes EV-FlowNet \cite{zhu2018ev}, STE-FlowNet \cite{ding2022spatio}, Spike-FlowNet \cite{lee2020spike} and Fusion-FlowNet \cite{lee2022fusion}. However, considering that intensity images suffer from motion blur and low dynamic range, using photometric consistency loss is prone to failure of supervision\cite{shiba2022secrets}. The other type of method uses event-based motion-compensation loss, including EV-FlowNet+ \cite{zhu2019unsupervised}, ConvGRU-EV-FlowNet \cite{hagenaars2021self}, ET-FlowNet \cite{tian2022event} and XLIF-EV-FlowNet \cite{hagenaars2021self}. They first use the predicted flows to warp the raw events for obtaining an image of warped events (IWE) and then measure the sharpness of the IWE for network training \cite{gallego2018unifying,gallego2019focus,zhang2022formulating}. However, there is a gap for existing motion-compensation losses in ensuing adequate geometric constraints for accurate event alignment.

\begin{figure*}[htbp]
\centering
\includegraphics[width=\textwidth]{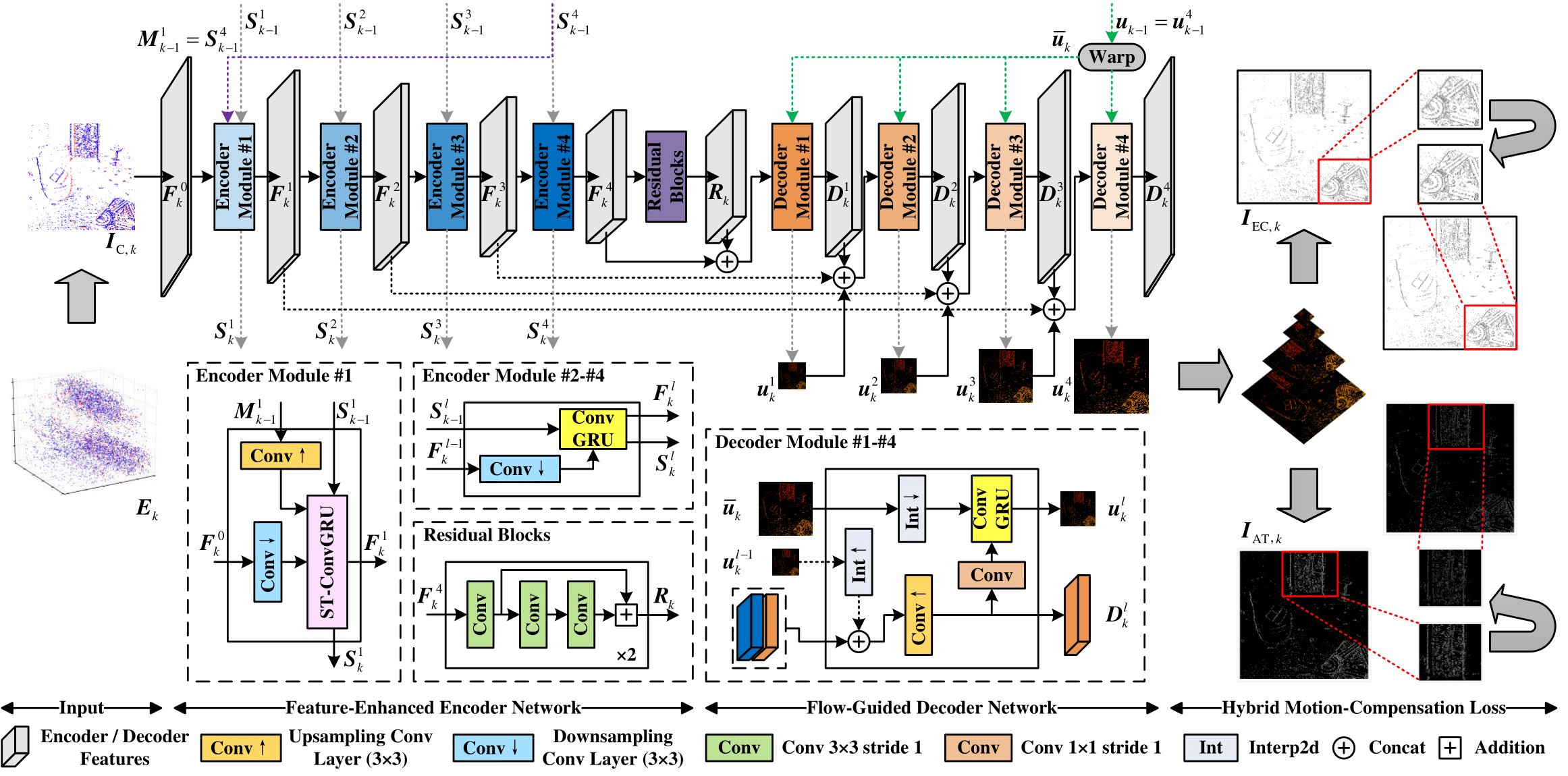}
\vspace{-2em}
\caption{Detailed illustration of the proposed EV-MGRFlowNet pipeline. Firstly, the event volume is converted into an event count image as input. Then, by introducing our proposed ST-ConvGRU, the feature-enhanced recurrent encoder network (FERE-Net) utilizes multi-level hidden states to obtain motion features. Next, the flow-guided decoder network (FGD-Net) predicts multi-scale optical flows combined with motion cues extracted from prior flows. Finally, the predicted flows are supervised by our proposed hybrid motion-compensation loss (HMC-Loss) and thereby produce motion-corrected average timestamp image and exponential count image.}
\label{fig:2}
\vspace{-1em}
\end{figure*}

\section{Methodology}
\label{sec:methodology}

\subsection{The Overall Architecture}
A detailed illustration of the proposed EV-MGRFlowNet pipeline is shown in Fig. \ref{fig:2}. Our EV-MGRFlowNet is a variant of a recurrent encoder-decoder network that comprises a feature-enhanced recurrent encoder network (FERE-Net) and a flow-guided decoder network (FGD-Net). The event volume is converted into an event count image and passed through FERE-Net for feature extraction. Then, FERE-Net incorporates ConvGRU and our proposed ST-ConvGRU to exploit multi-level hidden states for fully preserving motion patterns. Next, the output features from FERE-Net are fed into FGD-Net to realize multi-scale optical flow estimation guided by motion information extracted from prior flows. Finally, the estimated flows are supervised relying only on event data through our proposed hybrid motion-compensation loss (HMC-Loss), which measures the sharpness of both the average timestamp IWE and exponential count IWE.
\subsection{Event Camera and Event Representation}
In our scheme, we use event vision sensors (event cameras) to measure optical flow. The bio-inspired working principle of event cameras brings them unique advantages (e.g., high temporal resolution and high dynamic range). Different from frame-based cameras that record intensity images at a fixed rate, each pixel of the event camera can output events independently and asynchronously which are triggered by brightness changes. Specifically, when a pixel $\boldsymbol{x}_i=(x_i,y_i)^\top$ of the event camera satisfies the condition that the log intensity change exceeds the given contrast threshold $\vartheta$ at time $t_i$:
\begin{equation}
\begin{aligned}
\label{eq:1}
\left|\log\left(\frac{\boldsymbol{I}\left(\boldsymbol{x}_i, t_i\right)}{\boldsymbol{I}\left(\boldsymbol{x}_i,t_i-\Delta t_i\right)}\right)\right|\geq\vartheta,
\end{aligned}
\end{equation}
where $\boldsymbol{I}$ is the latent intensity on the pixel plane, $\Delta t_i$ is the time interval between two adjacent events at the same pixel, and an event $\boldsymbol{e}_i=(\boldsymbol{x}_i,t_i,p_i)^\top$ will be triggered at pixel $\boldsymbol{x}_i$ with the polarity $p_i\in\{+1,-1\}$. In this way, the sparse asynchronous event stream will be generated within a $x\text{-}y\text{-}t$ spatio-temporal coordinates. Since a single event carries little information alone, directly processing individual events with a deep network is difficult. Hence, it is necessary to aggregate events over short time windows to construct event representations that are amenable to deep network processing. First of all, we follow previous works and split the event stream into a series of event volumes $\boldsymbol{E}_k=\{\boldsymbol{e}_i\}_{i=1}^{\mathrm{N}_\text{e}}$ at fixed intervals. Then, the event volume needs to be converted into a suitable event representation. In this paper, we use the event count image $\boldsymbol{I}_{\text{C},k}\in\mathbb{R}^{2\times\mathrm{H}\times\mathrm{W}}$ as the input to our network, which is a widely used event representation \cite{hagenaars2021self, maqueda2018event, liu2023motion}. Specifically, it is expressed as follows:
\begin{equation}
\label{eq:2}
\begin{aligned}
\boldsymbol{I}_{\text{C},+,k}(x,y)&=\sum_{\boldsymbol{e}_i\in\boldsymbol{E}_{+,k}}p_i\boldsymbol{\delta}(x-x_i,y-y_i),\\
\boldsymbol{I}_{\text{C},-,k}(x,y)&=\sum_{\boldsymbol{e}_i\in\boldsymbol{E}_{-,k}}|p_i|\boldsymbol{\delta}(x-x_i,y-y_i),
\end{aligned}
\end{equation}
where $\boldsymbol{\delta}(\cdot)$ denotes the Dirac pulse. It can be seen that $\boldsymbol{I}_{\text{C},+,k}$ and $\boldsymbol{I}_{\text{C},-,k}$ are accumulated by the number of events triggered at each pixel in the positive events $\boldsymbol{E}_{+,k}$ and negative events $\boldsymbol{E}_{-,k}$, respectively.

\subsection{Feature-Enhanced Recurrent Encoder Network}
\begin{figure}[htbp]
\centering
\includegraphics[width=\columnwidth]{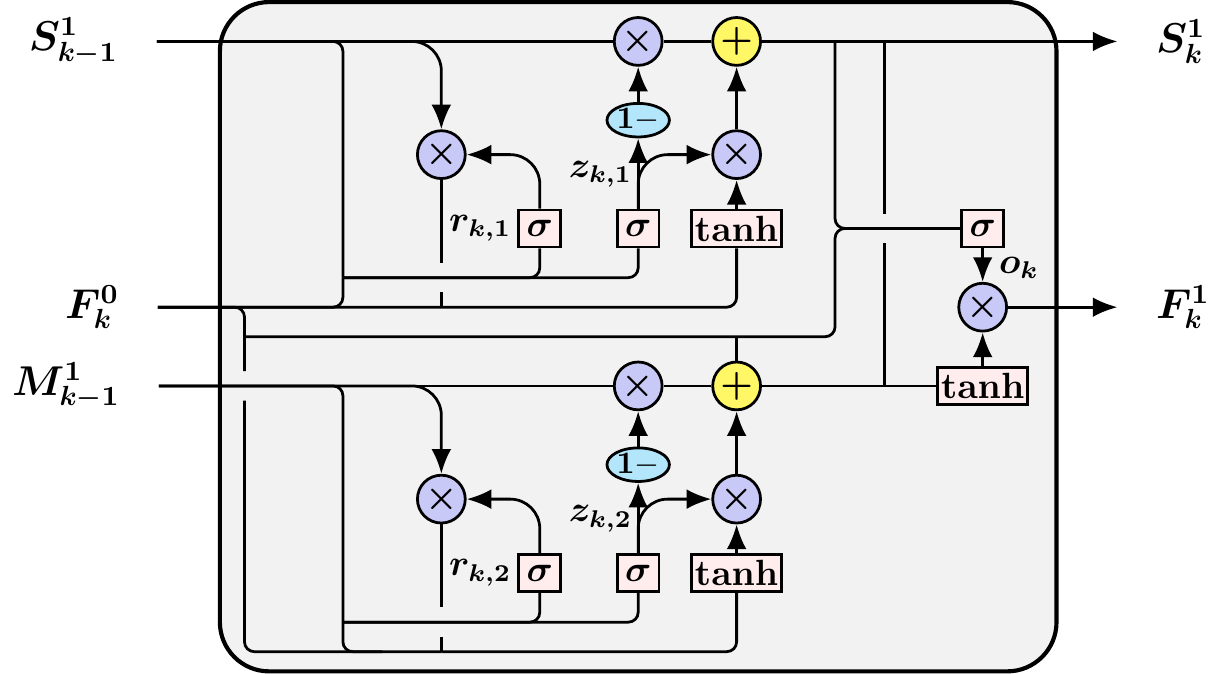}
\vspace{-2em}
\caption{Detailed illustration of the ST-ConvGRU network structure is as follows. The input event image $\boldsymbol{F}_{k}^{0}$, the hidden states $\boldsymbol{S}_{k-1}^1$ and $\boldsymbol{M}_{k-1}^1$ at the previous timestep as input. Another set of gate structures is adopted to memorize the information from $\boldsymbol{M}_{k-1}^1$  while retaining the standard ConvGRU for updating $\boldsymbol{S}_{k-1}^1$. The final output $\boldsymbol{F}_k^1$ and the hidden state $\boldsymbol{S}_k^1$ are obtained by fusing the two memorized information. Here, $\sigma$ denotes the sigmoid function. $\otimes$ and $\oplus$ represent pixel-wise multiplication and pixel-wise addition, respectively.}
\label{fig:3}
\vspace{-1em}
\end{figure}

After converting $\boldsymbol{E}_k$ into $\boldsymbol{I}_{\text{C},k}$, we can pass them into the encoder network for feature extraction. Introducing the convolutional gated recurrent unit (ConvGRU) \cite{ballas2015delving} into the encoder network is critical for extracting spatio-temporal features \cite{ding2022spatio, hidalgo2020learning, rebecq2019high,gehrig2021combining}. This is because ConvGRU can utilize the previous hidden state $\boldsymbol{H}_{k-1}$ and the current input $\boldsymbol{X}_{k}$ to enhance the current state $\boldsymbol{H}_{k}$, thereby explicitly encoding spatial information and capturing temporal dependencies:
\begin{equation}
\begin{aligned}
\label{eq:3}
&\boldsymbol{H}_{k} =\boldsymbol{f}_\text{ConvGRU}(\boldsymbol{X}_{k},\boldsymbol{H}_{k-1}),    
\end{aligned}       
\end{equation}
where $k$ denotes time step, and the structure of $\boldsymbol{f}_\text{ConvGRU}(\cdot)$ is as follows in detail: 
\begin{equation}
\label{eq:4}
\begin{aligned}
\boldsymbol{r}_{k} &=\boldsymbol{f}_\text{sigmoid} (\boldsymbol{W}_{\text{xr}} \ast \boldsymbol{X}_{k}+\boldsymbol{W}_{\text{hr}}\ast \boldsymbol{H}_{k-1}),   \\
\boldsymbol{z}_{k} &=\boldsymbol{f}_\text{sigmoid} (\boldsymbol{W}_{\text{xz}} \ast \boldsymbol{X}_{k} + \boldsymbol{W}_{\text{hz}} \ast \boldsymbol{H}_{k-1}),    \\
\boldsymbol{\hat{H}}_{k} &=\boldsymbol{f}_{\text{tanh}} \left( \boldsymbol{r}_{k} \odot \left(\boldsymbol{W}_{\text{hh}}\ast \boldsymbol{H}_{k-1}\right)+ \boldsymbol{W}_{\text{xh}} \ast \boldsymbol{X}_{k}\right),  \\
\boldsymbol{H}_{k} &= (1-\boldsymbol{z}_{k}) \odot \boldsymbol{H}_{k-1}+\boldsymbol{z}_{k} \odot  \boldsymbol{\hat{H}}_{k},    \\  
\end{aligned}       
\end{equation}
where $\boldsymbol{f}_\text{sigmoid}(\cdot)$ is the sigmoid activation function, and $\boldsymbol{f}_\text{tanh}(\cdot)$ is the hyperbolic tangent activation function. $\boldsymbol{W}_\text{xr}, \boldsymbol{W}_\text{hr}, \boldsymbol{W}_\text{xz}, \boldsymbol{W}_\text{hz}, \boldsymbol{W}_\text{hh}, \boldsymbol{W}_\text{xh}$ denote the learnable parameters, $\boldsymbol{r}_{k}$ and $\boldsymbol{z}_{k}$ are the reset gate and the update gate, respectively. $*$ and $\odot$ denote the convolution operator and the Hadamard product, respectively. The ConvGRU encoder network adopts the encoder-decoder RNN architecture that is widely used in event-based optical flow estimation works \cite{ding2022spatio, hagenaars2021self, tian2022event, gehrig2021combining}. For a 4-layer ConvGRU encoder network, event representations are encoded layer-by-layer, with hidden states being delivered from the first layer to the fourth one. However, the characteristic of stacked ConvGRUs determines that the memory cells of encoder modules at different resolutions will be updated only in the time domain. The low-level layer is then prone to discarding information from the high-level layer at the previous timestep, which records distinct spatio-temporal variations beneficial to capturing fine motion patterns \cite{wang2017predrnn}. In order to retain such significant spatio-temporal information, we propose a feature-enhanced recurrent encoder network (FERE-Net) into which a novel memory unit ST-ConvGRU is incorporated. As shown in Fig. \ref{fig:2}, FERE-Net consists of four layers of encoder modules $\{\boldsymbol{f}_{\text{E}_l}(\cdot)\}_{l=1}^4$. First, the input $\boldsymbol{I}_{\text{C},k}$ is passed into the first encoder module $\boldsymbol{f}_{\text{E}_1}(\cdot)$, specifically expressed as follows:
\begin{equation}
\label{eq:5}
\begin{aligned}
\boldsymbol{F}_{k}^{1},\boldsymbol{S}_{k}^{1}&=
\boldsymbol{f}_{\text{E}_1}(\boldsymbol{F}_{k}^{0},\boldsymbol{S}_{k-1}^{1},\boldsymbol{M}_{k-1}^{1})\\
&=\boldsymbol{f}_{\text{ST-ConvGRU}}\left(\boldsymbol{f}_{\text{conv}\downarrow}(\boldsymbol{F}_{k}^{0}),\boldsymbol{S}_{k-1}^{1},\boldsymbol{f}_{\text{conv}\uparrow}(\boldsymbol{M}_{k-1}^{1})\right),\\
\end{aligned}
\end{equation}
where $\boldsymbol{f}_{\text{conv}\downarrow}(\cdot)=\boldsymbol{f}_{\text{relu}}(\boldsymbol{f}_{\text{conv2d}}(\cdot))$ is the downsampling convolution layer, and $\boldsymbol{f}_{\text{conv}\uparrow}(\cdot)=\boldsymbol{f}_{\text{relu}}(\boldsymbol{f}_{\text{conv2d}}(\boldsymbol{f}_{\text{Interp2d}\uparrow}(\cdot)))$ is the upsampling convolution layer. 
Here, the input of $\boldsymbol{f}_{\text{E}_1}(\cdot)$ at timestep $k$ contains three parts: the input event representation $\boldsymbol{F}_{k}^{0}=\boldsymbol{I}_{\text{C},k}$, the previous hidden state $\boldsymbol{S}_{k-1}^{1}$ from the first encoder module and the previous hidden state $\boldsymbol{M}_{k-1}^{1}=\boldsymbol{S}_{k-1}^{4}$ from the fourth encoder module. Its purpose is to utilize previous features $\boldsymbol{M}_{k-1}^{1}$ and $\boldsymbol{S}_{k-1}^{1}$ to enrich the encoder network and output the updated feature maps $\boldsymbol{F}_{k}^{1}$ and the updated hidden state $\boldsymbol{S}_{k}^{1}$.
As we can see, the critical component to realize feature enrichment within the first encoder module $\boldsymbol{f}_{\text{E}_1}(\cdot)$ is our proposed ST-ConvGRU, which is shown in Fig. \ref{fig:3}. In contrast to ConvGRU, while retaining the standard ConvGRU for the update of the memory cell $\boldsymbol{S}$, it constructs another set of gate structures for a new memory cell $\boldsymbol{M}$. After obtaining the updated memory $\boldsymbol{S}_{k}^{1}$ and $\boldsymbol{\bar{M}}_k$ at the current timestep, we realize efficient memory fusion by applying an output gate $\boldsymbol{o}_k$ and a $1\times 1$ convolution layer. In more detail, ST-ConvGRU is defined by the following equations:
\begin{equation}
\label{eq:6}
\begin{aligned}
\boldsymbol{S}_{k}^{1} &=\boldsymbol{f}_{\text{ConvGRU}}(\boldsymbol{f}_{\text{conv}\downarrow}(\boldsymbol{F}_{k}^{0}), \boldsymbol{S}_{k-1}^{1}),\\
\boldsymbol{\bar{M}}_{k} &=\boldsymbol{f}_{\text{ConvGRU}}(\boldsymbol{f}_{\text{conv}\downarrow}(\boldsymbol{F}_{k}^{0}), \boldsymbol{M}_{k-1}^{1}),\\
\boldsymbol{o}_{k} &=\boldsymbol{f}_\text{sigmoid} (\boldsymbol{W}_{\text{fo}} \ast \boldsymbol{f}_{\text{conv}\downarrow}(\boldsymbol{F}_{k}^{0}) + \boldsymbol{W}_{\text{so}} \ast \boldsymbol{S}_{k}^{1}+\boldsymbol{W}_{\text{mo}} \ast \boldsymbol{\bar{M}}_{k}),\\
\boldsymbol{F}_{k}^{1} &=\boldsymbol{o}_{k} \ast \boldsymbol{f}_{\text{tanh}}(\boldsymbol{W}_{\text{mm}}\ast \boldsymbol{\bar{M}}_{k}+ \boldsymbol{W}_{\text{ss}} \ast \boldsymbol{S}_{k}^{1}),\\    
\end{aligned}       
\end{equation}
where $\boldsymbol{W}_\text{fo}$, $\boldsymbol{W}_\text{so}$, $\boldsymbol{W}_\text{mo}$, $\boldsymbol{W}_\text{mm}$ and $\boldsymbol{W}_\text{ss}$ are the trainable parameters. Then, the feature maps  $\boldsymbol{F}_{k}^{1}$ go through the following encoder modules $\boldsymbol{f}_{\text{E}_2}(\cdot)$, $\boldsymbol{f}_{\text{E}_3}(\cdot)$ and $\boldsymbol{f}_{\text{E}_4}(\cdot)$ sequentially for feature extraction:
\begin{equation}
\label{eq:7}
\begin{aligned}
\boldsymbol{F}_{k}^{2},\boldsymbol{S}_{k}^{2}&=
\boldsymbol{f}_{\text{E}_2}(\boldsymbol{F}_{k}^{1},\boldsymbol{S}_{k-1}^{2})\\
&=\boldsymbol{f}_{\text{ConvGRU}}(\boldsymbol{f}_{\text{conv}\downarrow}(\boldsymbol{F}_{k}^{1}),\boldsymbol{S}_{k-1}^{2}),\\
\boldsymbol{F}_{k}^{3},\boldsymbol{S}_{k}^{3}&=
\boldsymbol{f}_{\text{E}_3}(\boldsymbol{F}_{k}^{2},\boldsymbol{S}_{k-1}^{3})\\
&=\boldsymbol{f}_{\text{ConvGRU}}(\boldsymbol{f}_{\text{conv}\downarrow}(\boldsymbol{F}_{k}^{2}),\boldsymbol{S}_{k-1}^{3}),\\
\boldsymbol{F}_{k}^{4},\boldsymbol{S}_{k}^{4}&=
\boldsymbol{f}_{\text{E}_4}(\boldsymbol{F}_{k}^{4},\boldsymbol{S}_{k-1}^{4})\\
&=\boldsymbol{f}_{\text{ConvGRU}}(\boldsymbol{f}_{\text{conv}\downarrow}(\boldsymbol{F}_{k}^{4}),\boldsymbol{S}_{k-1}^{4}),\\
\end{aligned}
\end{equation}
where the hidden states of each encoder module at the previous timestep are $\boldsymbol{S}_{k-1}^{2}$, $\boldsymbol{S}_{k-1}^{3}$ and $\boldsymbol{S}_{k-1}^{4}$ respectively. The updated feature maps and hidden states through each encoder module are $\boldsymbol{F}_{k}^{2}$, $\boldsymbol{F}_{k}^{3}$, $\boldsymbol{F}_{k}^{4}$ and $\boldsymbol{S}_{k}^{2}$, $\boldsymbol{S}_{k}^{3}$, $\boldsymbol{S}_{k}^{4}$ respectively.
Note that the memory unit of the first encoder module is ST-ConvGRU, while the remaining encoder modules use ConvGRU as the memory unit for the fusion of key features $\boldsymbol{M}_{k-1}^1$. Finally, as shown in Fig. \ref{fig:2}, the output feature maps $\boldsymbol{F}_{k}^{4}$ from the encoder network are fed into a residual module for further feature extraction:
\begin{equation}
\label{eq:8}
\begin{aligned}
\boldsymbol{R}_{k} &= \boldsymbol{f}_{\text{RB}_2}(\boldsymbol{f}_{\text{RB}_1}(\boldsymbol{F}_{k}^{4})).   \\
\end{aligned}
\end{equation}
The network consists of two residual blocks $\boldsymbol{f}_{\text{RB}_1}(\cdot), \boldsymbol{f}_{\text{RB}_2}(\cdot)$ to output the final encoder results $\boldsymbol{R}_{k}$.

\subsection{Flow-Guided Decoder Network}
Previous optical flow estimation networks based on the U-Net architecture \cite{zhu2018ev, zhu2019unsupervised, hagenaars2021self, ding2022spatio} have rarely been able to utilize prior optical flows. However, they can provide effective initialization and supplementary information for the current optical flow estimation, which has been validated in \cite{gehrig2021raft} based on the RAFT \cite{teed2020raft} architecture. As shown in Fig. \ref{fig:2}, we propose a flow-guided decoder network (FGD-Net) to utilize prior optical flows for current optical flow estimation. First, we need to align consecutive optical flow estimations $\boldsymbol{u}_{k-1}$ and $\boldsymbol{u}_{k}$ through forward warping since their source pixel locations are inconsistent. Given that the constant flow assumption, we can obain the warped prior flow $\boldsymbol{\bar{u}}_{k}$ from $\boldsymbol{u}_{k-1}$. The transformation equation is shown below:
\begin{equation}
\label{eq:9}
\begin{aligned}
\boldsymbol{\bar{u}}_{k}(x+\mathbf{u}_{k-1}(x,y),y+\mathbf{v}_{k-1}(x,y))=\boldsymbol{u}_{k-1}(x,y),
\end{aligned}
\end{equation}
where $\boldsymbol{\bar{u}}_{k}=[\mathbf{\bar{u}}_{k},\mathbf{\bar{v}}_{k}]^\top$. 
Then, the entire decoder network consists of four decoder modules $\{\boldsymbol{f}_{\text{D}_l}(\cdot)\}_{l=1}^4$ to achieve multi-scale optical flow estimation, and the entire decoding process is as follows: 
\begin{equation}
\label{eq:10}
\begin{aligned}
\boldsymbol{D}_{k}^{1},\boldsymbol{u}_{k}^1&=
\boldsymbol{f}_{\text{D}_1}(\boldsymbol{R}_{k},\boldsymbol{F}_{k}^{4},\boldsymbol{\bar{u}}_{k}),\\
\boldsymbol{D}_{k}^{2},\boldsymbol{u}_{k}^2&=
\boldsymbol{f}_{\text{D}_2}(\boldsymbol{D}_{k}^{1},\boldsymbol{F}_{k}^{3},\boldsymbol{\bar{u}}_{k},\boldsymbol{u}_{k}^1),\\
\boldsymbol{D}_{k}^{3},\boldsymbol{u}_{k}^3&=
\boldsymbol{f}_{\text{D}_3}(\boldsymbol{D}_{k}^{2},\boldsymbol{F}_{k}^{2},\boldsymbol{\bar{u}}_{k},\boldsymbol{u}_{k}^2),\\
\boldsymbol{D}_{k}^{4},\boldsymbol{u}_{k}^4&=
\boldsymbol{f}_{\text{D}_4}(\boldsymbol{D}_{k}^{3},\boldsymbol{F}_{k}^{1},\boldsymbol{\bar{u}}_{k},\boldsymbol{u}_{k}^3),\\
\end{aligned}
\end{equation}
where $\{\boldsymbol{F}_k^l\}_{l=1}^4$ and $\boldsymbol{R}_{k}$ are the multi-scale feature maps and the residual feature maps obtained from FERE-Net, respectively. $\boldsymbol{D}_{k}^{1}$, $\boldsymbol{D}_{k}^{2}$, $\boldsymbol{D}_{k}^{3}$ and $\boldsymbol{D}_{k}^4$ are the intermediate feature maps. $\boldsymbol{u}_{k}^1$, $\boldsymbol{u}_{k}^2$, $\boldsymbol{u}_{k}^3$ and $\boldsymbol{u}_{k}^4$ are multi-scale estimated optical flows. When passing through the decoder module $\boldsymbol{f}_{\text{D}_l}(\cdot)$, the last layer's feature maps $\boldsymbol{D}_{k}^{l-1}$, the corresponding encoder features $\boldsymbol{F}_{k}^{5-l}$ and the warped prior flow $\boldsymbol{\bar{u}}_{k}$ are served as input. They are then fed into $\boldsymbol{f}_{\text{D}_l}(\cdot)$ to generate the final predicted flow $\boldsymbol{{u}}_{k}^{l}$ and the intermediate feature maps $\boldsymbol{D}_{k}^{l}$. The details of each decoder module $\boldsymbol{f}_{\text{D}_l}(\cdot)$ are defined by the following equations:
\begin{equation}
\label{eq:11}
\begin{aligned}
\boldsymbol{D}_{k}^{l}&=\left\{
\begin{array}{ll}
\boldsymbol{f}_{\text{conv}\uparrow}(\boldsymbol{R}_{k}\oplus\boldsymbol{F}_{k}^{5-l})   &l=1\\
\boldsymbol{f}_{\text{conv}\uparrow}(\boldsymbol{D}_{k}^{l-1}\oplus\boldsymbol{F}_{k}^{5-l}\oplus \boldsymbol{f}_{\text{Interp2d}\uparrow}(\boldsymbol{u}_{k}^{l-1}))    &l\neq1\\
\end{array}
\right.,\\
\boldsymbol{\hat{u}}_{k}^{l}&=\boldsymbol{f}_{\text{tanh}}(\boldsymbol{f}_{\text{conv2d}}(\boldsymbol{D}_{k}^{l})),\\
\boldsymbol{\bar{u}}_{k}^{l}&=\boldsymbol{f}_{\text{Interp2d}\downarrow}( \boldsymbol{\bar{u}}_{k}),\\
\boldsymbol{u}_{k}^{l}&=\boldsymbol{f}_{\text{ConvGRU}}(\boldsymbol{\hat{u}}_{k}^{l},\boldsymbol{\bar{u}}_{k}^{l}),\\
\end{aligned}
\end{equation}
where $\oplus$ denotes concatenation along the channel. In more detail, $\boldsymbol{D}_{k}^{l-1}$, $\boldsymbol{F}_k^{5-l}$ and the upsampled  $\boldsymbol{u}_{k}^{l-1}$ are first concatenated to generate the initial predicted flow $\boldsymbol{\hat{u}}_{k}^{l}$ and the intermediate feature maps $\boldsymbol{D}_{k}^{l}$. Then, we utilize ConvGRU to extract motion information beneficial for $\boldsymbol{\hat{u}}_{k}^{l}$ from $\boldsymbol{\bar{u}}_{k}$, resulting in refined result $\boldsymbol{u}_{k}^{l}$. Note that $\boldsymbol{\bar{u}}_{k}$ is required to be downsampled to match the resolution of $\boldsymbol{\hat{u}}_{k}^{l}$ by $\boldsymbol{f}_{\text{Interp2d}\downarrow}(\cdot)$. In this way, we can serve $\boldsymbol{u}_{k}^4$ as the current estimated flow $\boldsymbol{u}_{k}$. Meanwhile, $\boldsymbol{u}_{k}$ is leveraged as the prior flow to guide generation of next predicted optical flow.

\subsection{Hybrid Motion-Compensation Loss}
For network training, we propose a novel hybrid motion-compensation loss (HMC-Loss) to train EV-MGRFlowNet with unsupervised learning. In recent years, several works \cite{zhu2019unsupervised, hagenaars2021self, tian2022event} have introduced motion compensation to design losses relying solely on event data for unsupervised learning. Specifically, they first use the predicted flows $\boldsymbol{u}_k\in\mathbb{R}^{2\times\mathrm{H}\times\mathrm{W}}$ from the network $\boldsymbol{f}_{\text{flownet}}(\cdot)$ to propagate the raw event volume $\boldsymbol{E}_k$ to a reference time $t^{\prime}$ for obtaining the warped event volume $\boldsymbol{E}_k^{\prime}$:
\begin{equation}
\label{eq:12}
\begin{aligned}
\boldsymbol{u}_k&= \boldsymbol{f}_{\text{flownet}}(\boldsymbol{E}_k),\\
\boldsymbol{E}_k^{\prime}&=\{\boldsymbol{e}_i^{\prime}\}_{i=1}^{\mathrm{N}_\text{e}}\\
&=\{(\boldsymbol{x}_i^{\prime},t_i,p_i)^\top\}_{i=1}^{\mathrm{N}_\text{e}}\\
&=\{(\boldsymbol{x}_i+(t^{\prime}-t_i) \cdot \boldsymbol{u}_k( \boldsymbol{x}_i),t_i,p_i)^\top\}_{i=1}^{\mathrm{N}_\text{e}},\\
\end{aligned}
\end{equation}
where $\boldsymbol{u}_k$ means the per-pixel optical flow. Then, a motion-compensation loss is designed to measure the sharpness of the IWE generated by $\boldsymbol{E}_k^{\prime}$. At present, the widely used motion-compensation loss $L_\text{AT}$ 
measures the sharpness of the average timestamp IWE $\boldsymbol{I}_{\text{AT}}\in\mathbb{R}^{2\times\mathrm{H}\times\mathrm{W}}$ \cite{zhu2019unsupervised}. Specifically, $\boldsymbol{I}_{\text{AT}}$ is generated as follows:
\begin{equation}
\label{eq:13}
\begin{aligned}
\boldsymbol{I}_{\text{AT},+,k}(\boldsymbol{x};\boldsymbol{u}_k|t^{\prime})=\frac{\sum_{\boldsymbol{e}_i^{\prime}\in\boldsymbol{E}_{+,k}^{\prime}}{\kappa(x-x_i^{\prime},y-y_i^{\prime})t_i}}{\sum_{\boldsymbol{e}_i^{\prime}\in\boldsymbol{E}_{+,k}^{\prime}}{\kappa(x-x_i^{\prime},y-y_i^{\prime})}},\\
\boldsymbol{I}_{\text{AT},-,k}(\boldsymbol{x};\boldsymbol{u}_k|t^{\prime})=\frac{\sum_{\boldsymbol{e}_i^{\prime}\in\boldsymbol{E}_{-,k}^{\prime}}{\kappa(x-x_i^{\prime},y-y_i^{\prime})t_i}}{\sum_{\boldsymbol{e}_i^{\prime}\in \boldsymbol{E}_{-,k}^{\prime}}{\kappa(x-x_i^{\prime},y-y_i^{\prime})}},\\
\end{aligned}
\end{equation}
where $\kappa$ is the bilinear sampling kernel, $\kappa(\Delta x,\Delta y)=\text{max}(0,1-|\Delta x|)\cdot\text{max}(0,1-|\Delta y|)$. Then $L_\text{AT}$ minimizes the sum of squares of $\boldsymbol{I}_{\text{AT}}$. Here, the reformulation of $L_\text{AT}$ at $t^{\prime}$ is as follows:
\begin{equation}
\label{eq:14}
\begin{aligned}
&L_{\text{AT}}(\boldsymbol{u}_k|t^{\prime})\\
&\quad=\sum_{\boldsymbol{x}}\boldsymbol{I}_{\text{AT},+,k}(\boldsymbol{x};\boldsymbol{u}_k|t^{\prime})^2+\sum_{\boldsymbol{x}}\boldsymbol{I}_{\text{AT},-,k}(\boldsymbol{x};\boldsymbol{u}_k|t^{\prime})^2.
\end{aligned}
\end{equation}
We choose $L_\text{AT}$ as part of our HMC-Loss since it has excellent accuracy performance among deep learning works. However, $L_\text{AT}$ still lacks sufficient geometric constraints for accurate event alignment, consequently causing the trained network to produce undesired flows \cite{stoffregen2019event,shiba2022event}. Hence, inspired by \cite{stoffregen2019event}, we adopt the exponential count IWE $\boldsymbol{I}_{\text{EC}}\in\mathbb{R}^{2\times\mathrm{H}\times\mathrm{W}}$ to design a novel motion-compensation loss $L_{\text{EC}}$ that imposes stronger geometric constraints on the event data. Specifically, $\boldsymbol{I}_{\text{EC}}$ is generated as follows:
\begin{equation}
\label{eq:15}
\begin{aligned}
\boldsymbol{I}_{\text{EC},+,k}(\boldsymbol{x};\boldsymbol{u}_k|t^{\prime})= \text{exp}(-\alpha\sum_{\boldsymbol{e_i}^{\prime}\in \boldsymbol{E}_{+,k}^{\prime}}{\kappa(x-x_i^{\prime},y-y_i^{\prime})}),   \\
\boldsymbol{I}_{\text{EC},-,k}(\boldsymbol{x};\boldsymbol{u}_k|t^{\prime})= \text{exp}(-\alpha\sum_{\boldsymbol{e_i}^{\prime}\in \boldsymbol{E}_{-,k}^{\prime}}{\kappa(x-x_i^{\prime},y-y_i^{\prime})}),   \\
\end{aligned}
\end{equation}
where $\alpha$ is the saturation factor. Then we design a novel motion-compensation loss $L_{\text{EC}}$ to penalize the event dispersion effect on the $\boldsymbol{I}_{\text{EC}}$:
\begin{equation}
\label{eq:16}
\begin{aligned}
&L_{\text{EC}}(\boldsymbol{u}_k|t^{\prime})\\
&\quad=\frac{\mathrm{N}}{\sum_{\boldsymbol{x}}\boldsymbol{I}_{\text{EC},+,k}(\boldsymbol{x};\boldsymbol{u}_k|t^{\prime})}+\frac{\mathrm{N}}{\sum_{\boldsymbol{x}}\boldsymbol{I}_{\text{EC},-,k}(\boldsymbol{x};\boldsymbol{u}_k|t^{\prime})}-2 ,
\end{aligned}
\end{equation}
where $\mathrm{N}=\mathrm{H}\times\mathrm{W}$ is the total number of pixels. Compared with $L_\text{AT}$, $L_{\text{EC}}$ focuses on the pixels with no event rather than the pixels with at least one event. Moreover, $L_{\text{EC}}$ places a relatively greater focus on the global sharpness of IWE compared to $L_{\text{AT}}$. Therefore, combining $L_\text{AT}$ with $L_{\text{EC}}$ can further strengthen geometric constraints for event alignment. Accordingly, we name our proposed loss $L_\text{HMC}$ the hybrid motion-compensation loss. Next, we follow \cite{zhu2019unsupervised} and compute the loss at $t_1$ and $t_{\mathrm{N}_\text{e}}$. Furthermore, we require a smoothing loss $L_\text{smooth}$ to minimize the flow difference between neighboring pixels. Finally, the total loss $L_\text{HMC}$ for training the network is defined as follows:
\begin{equation}
\label{eq:17}
\begin{aligned}
L_{\text{HMC}}(\boldsymbol{u}_k)&=\left(L_{\text{AT}}(\boldsymbol{u}_k|t_{\mathrm{1}})+L_{\text{AT}}(\boldsymbol{u}_k|t_{\mathrm{N}_\text{e}})\right)\\
&\quad+\lambda_1\left(L_{\text{EC}}(\boldsymbol{u}_k|t_{\mathrm{1}})+L_{\text{EC}}(\boldsymbol{u}_k|t_{\mathrm{N}_\text{e}})\right)\\
&\quad+\lambda_2 L_\text{smooth}(\boldsymbol{u}_k),   \\
\end{aligned}
\end{equation}
where $\lambda_1$ and $\lambda_2$ are scalars balancing the effect of the three losses. Moreover, $L_{\text{smooth}}$ follows the Charbonnier smoothness prior from \cite{hagenaars2021self}.

\begin{table*}[htbp]
\begin{center}
\caption{Comparison of our EV-MGRFlowNet against SOTA unsupervised learning methods \cite{zhu2018ev, zhu2019unsupervised, lee2020spike, hagenaars2021self, ding2022spatio, tian2022event} on MVSEC \cite{zhu2018multivehicle} in the case of $\mathrm{d}t=1$ and $\mathrm{d}t=4$. Best in bold, runner-up underlined. $\text{USL}_{\text{I}}$: using photometric consistency loss between two consecutive intensity images, $\text{USL}_{\text{E}}$: using event-based moton-compensation loss from event data.}

\newcommand{\tabincell}[2]{\begin{tabular}{@{}#1@{}}#2\end{tabular}}
\setlength{\tabcolsep}{0.005\linewidth}
\begin{tabular}{ccccccccccc p{1cm}}
\toprule
\multirow{3}{*}{\tabincell{c}{Frame\\Interval}}
&\multirow{3}{*}{\tabincell{c}{Learning}}
&\multirow{3}{*}{\tabincell{c}{Methods}}
&\multicolumn{8}{c}{\tabincell{c}{MVSEC\cite{zhu2018multivehicle} Dataset}}   \\
\cline{4-11}
&   &
&\multicolumn{2}{c}{\tabincell{c}{outdoor\_day1}}
&\multicolumn{2}{c}{\tabincell{c}{indoor\_flying1}}
&\multicolumn{2}{c}{\tabincell{c}{indoor\_flying2}}
&\multicolumn{2}{c}{\tabincell{c}{indoor\_flying3}}\\
\cline{4-11}
& & &AEE$\downarrow$ 
&Outlier (\%)$\downarrow$ 
&AEE$\downarrow$ 
&Outlier (\%)$\downarrow$ 
&AEE$\downarrow$ 
&Outlier (\%)$\downarrow$ 
&AEE$\downarrow$ 
&Outlier (\%)$\downarrow$  \\
\midrule
\multirow{8}{*}{\tabincell{c}{$\mathrm{d}t=1$}}
&\multirow{3}{*}{\tabincell{c}{$\text{USL}_{\text{I}}$}} &EV-FlowNet\cite{zhu2018ev}  & 0.49  & 0.20  & 1.03  & 2.20   & 1.72   & 15.10 & 1.53  & 11.90     \\
\cline{3-11}
&  & Spike-FlowNet\cite{lee2020spike}  & 0.49  & --- & 0.84   & ---    & 1.28  & ---   & 1.11  & ---  \\
\cline{3-11}
&  & STE-FlowNet\cite{ding2022spatio}  & 0.42  & \textbf{0.0}  & \underline{0.57}  & \underline{0.10}  & \underline{0.79} & \textbf{1.60}  & \underline{0.72}  & \underline{1.30}   \\
\cline{2-11}
&\multirow{5}{*}{\tabincell{c}{$\text{USL}_{\text{E}}$}}   & EV-FlowNet+\cite{zhu2019unsupervised}  & \underline{0.32} & \textbf{0.0} & 0.58 & \textbf{0.0}  & 1.02  & 4.00 & 0.87 & 3.00   \\
\cline{3-11}
&   & XLIF-EV-FlowNet \cite{hagenaars2021self}  & 0.45 & 0.16  & 0.73  & 0.92   & 1.45  & 12.18   & 1.17  & 8.35  \\
\cline{3-11}
&   & ConvGRU-EV-FlowNet \cite{hagenaars2021self}  & 0.47   & 0.25 & 0.60 & 0.51& 1.17 & 8.06 & 0.93   & 5.64    \\
\cline{3-11}
&   & ET-FlowNet\cite{tian2022event}  & 0.39& 0.12 & \underline{0.57} & 0.53  & 1.20 & 8.48  & 0.95  & 5.73   \\
\cline{3-11}
&   &  EV-MGRFlowNet (Ours)  & \textbf{0.28}  & \underline{0.02}  & \textbf{0.41}  & 0.17  & \textbf{0.70} & \underline{2.35}  & \textbf{0.59}  & \textbf{1.29} \\
\midrule
\multirow{8}{*}{\tabincell{c}{$\mathrm{d}t=4$}}
&\multirow{3}{*}{\tabincell{c}{$\text{USL}_{\text{I}}$}}  & EV-FlowNet\cite{zhu2018ev}  & 1.23  & 7.30 & 2.25    & 24.70 & 4.05    & 45.30  & 3.45    & 39.70     \\
\cline{3-11}
&   & Spike-FlowNet\cite{lee2020spike}  & \underline{1.09}   & ---   & 2.24  & ---   & 3.83  & ---   & 3.18  & ---   \\
\cline{3-11}
&  & STE-FlowNet\cite{ding2022spatio}  & \textbf{0.99}    & \textbf{3.90}     & \underline{1.77}       & \underline{14.70}             & \underline{2.52}     & \underline{26.10}        & \underline{2.23}      &\underline{22.10}  \\
\cline{2-11}
&\multirow{5}{*}{\tabincell{c}{$\text{USL}_{\text{E}}$}}    & EV-FlowNet+\cite{zhu2019unsupervised}   & 1.30    & 9.70   & 2.18   & 24.20     & 3.85   & 46.80 & 3.18    & 47.80  \\
\cline{3-11}
&   & XLIF-EV-FlowNet \cite{hagenaars2021self}    & 1.67   & 12.69   & 2.72   & 31.69    & 4.93   & 51.36    & 3.91    & 42.52    \\
\cline{3-11}
&   & ConvGRU-EV-FlowNet \cite{hagenaars2021self}  & 1.69  & 12.50  & 2.16   & 21.51  & 3.90  & 40.72  & 3.00  & 29.60  \\
\cline{3-11}
&   & ET-FlowNet\cite{tian2022event}   & 1.47  & 9.17   & 2.08   & 20.02 & 3.99  & 41.33  & 3.13  & 31.70  \\
\cline{3-11}
&   & EV-MGRFlowNet (Ours)      & 1.10   &\underline{6.22}    & {\textbf{1.50}}   & {\textbf{8.67}}  & {\textbf{2.39}}     & {\textbf{23.70}} & {\textbf{2.06}}   & {\textbf{18.00}}  \\

\bottomrule
\end{tabular}
\label{tab:1}
\vspace{-1.5em}
\end{center}
\end{table*}

\section{Experiments}
\label{sec:experiments}
\subsection{Experimental Setup}

\subsubsection{Dataset Selection}
For fair comparisons, we use the MVSEC \cite{zhu2018multivehicle} dataset for evaluation, which is now the de facto benchmark for event-based optical flow estimation. In more detail, we evaluate networks on the same test sequences as existing works \cite{zhu2018ev,zhu2019unsupervised,lee2020spike,ding2022spatio}, including indoor\_flying1, indoor\_flying2, indoor\_flying3 and a specific fragment of outdoor\_day1. For training, early works \cite{zhu2018ev,zhu2019unsupervised} directly used outdoor\_day2 from MVSEC as the training dataset. However, under this training configuration, limited and biased training data can lead to overfitting. By contrast, the UZH-FPV drone racing dataset \cite{delmerico2019we} has a wider distribution of optical flow compared to MVSEC. Therefore, we follow the training configuration from recent works \cite{hagenaars2021self, tian2022event} and train our EV-MGRFlowNet on the forward-facing sequences from UZH-FPV.
\begin{figure}[htbp]
\centering
\includegraphics[width=\columnwidth]{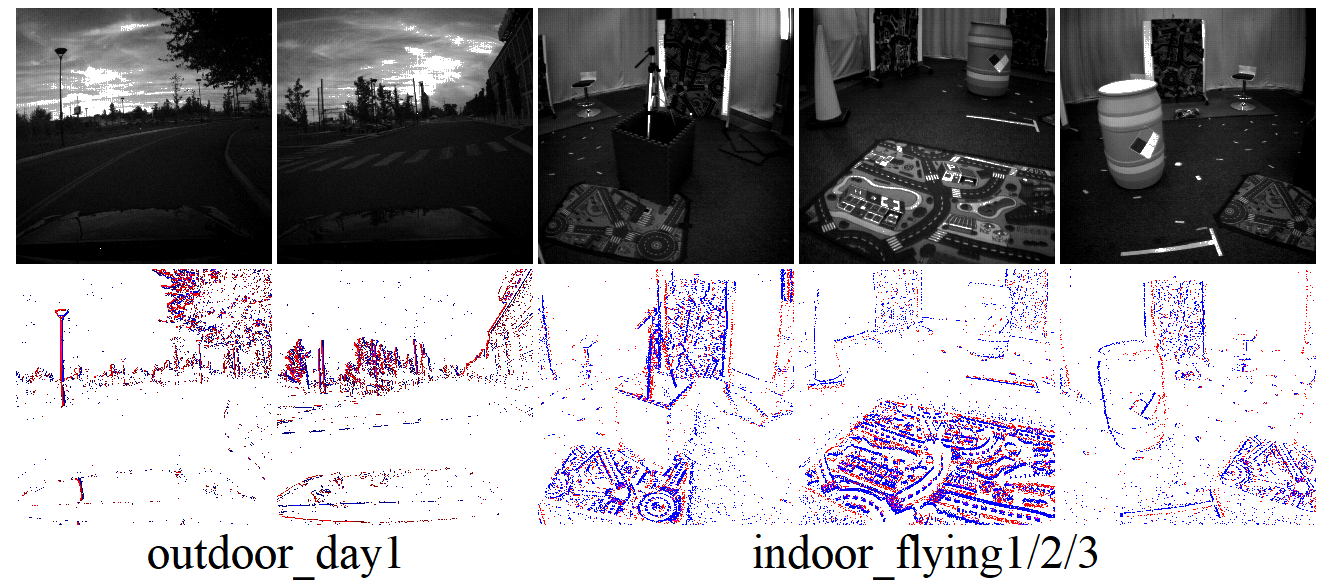}
\vspace{-2em}
\caption{Scenarios of MVSEC \cite{zhu2018multivehicle} dataset in our experiments, including outdoor\_day1, indoor\_day1, indoor\_day2 and indoor\_day3.}
\label{fig:4}
\vspace{-1em}
\end{figure}

\subsubsection{Parameters in Training}
To learn spatio-temporal association between events, we train our EV-MGRFlowNet by sequential learning \cite{hagenaars2021self} with many groups of event sequences. Each sequence of events contains $L$ consecutive non-overlapping event volumes (Sequence length $L=10$). Then, the proposed HMC-Loss is used for supervision. We use the ADAM optimizer with a learning rate of 0.0001 for 100 epochs. Here, we empirically set the saturation factor $\alpha=0.6$ for $L_{\text{EC}}$ and the balanced weights $\lambda_1 =1$ and $\lambda_2 = 0.001$ for $L_{\text{HMC}}$, respectively.

\subsubsection{Evaluation Metrics}
We refer to previous works \cite{zhu2018ev, zhu2019unsupervised, lee2020spike, hagenaars2021self, ding2022spatio, tian2022event} and use the following evaluation metrics: average endpoint error (AEE) and outlier rate (denoted as Outlier (\%)). Note that we only consider the pixels with valid ground truth flow and at least one event when calculating AEE. Outlier (\%) reports the percentage of points with endpoint error greater than 3 pixels and 5\% of the flow vector magnitude. In addition, previous works have used two frame intervals for optical flow evaluation, including optical flow between two adjacent frames ($\mathrm{d}t=1$) and optical flow spaced three frames apart ($\mathrm{d}t=4$), respectively.

\begin{figure*}[htbp]
\centering
\subfigure[]{\label{fig:4a}\includegraphics[width=0.141\textwidth]{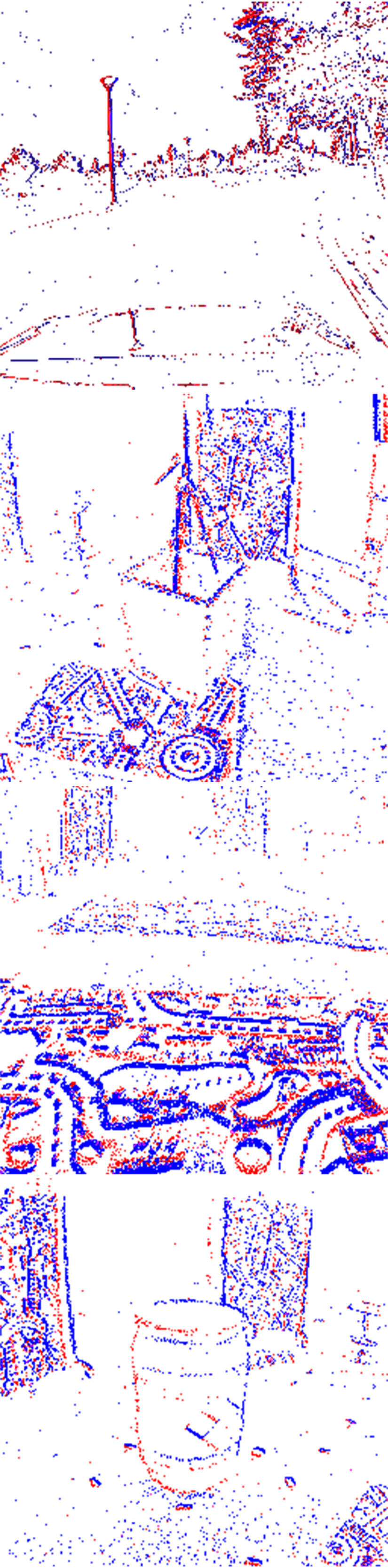}}
\hspace{-0.6em}
\subfigure[]{\label{fig:4b}\includegraphics[width=0.141\textwidth]{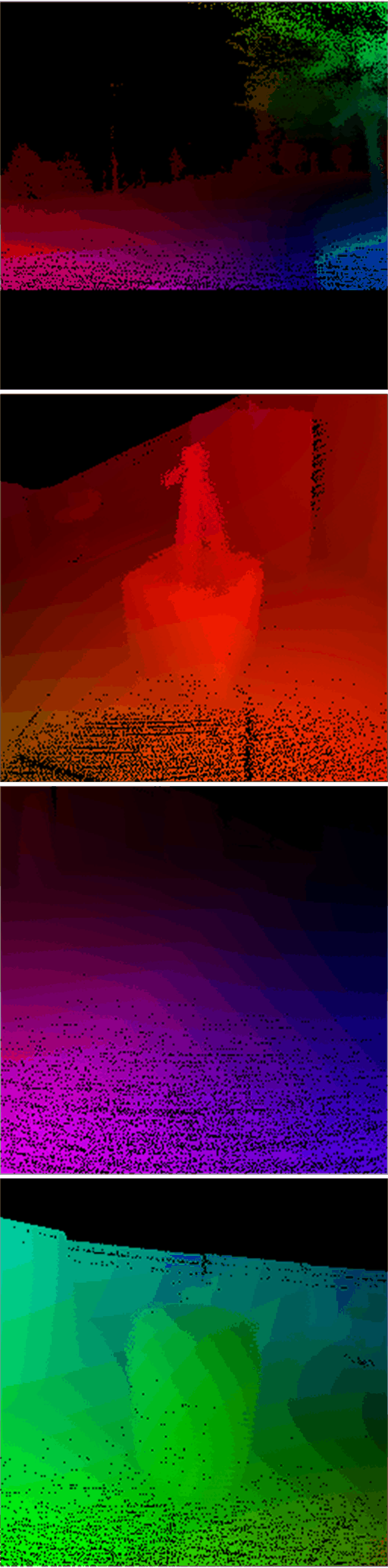}}
\hspace{-0.6em}
\subfigure[]{\label{fig:4c}\includegraphics[width=0.141\textwidth]{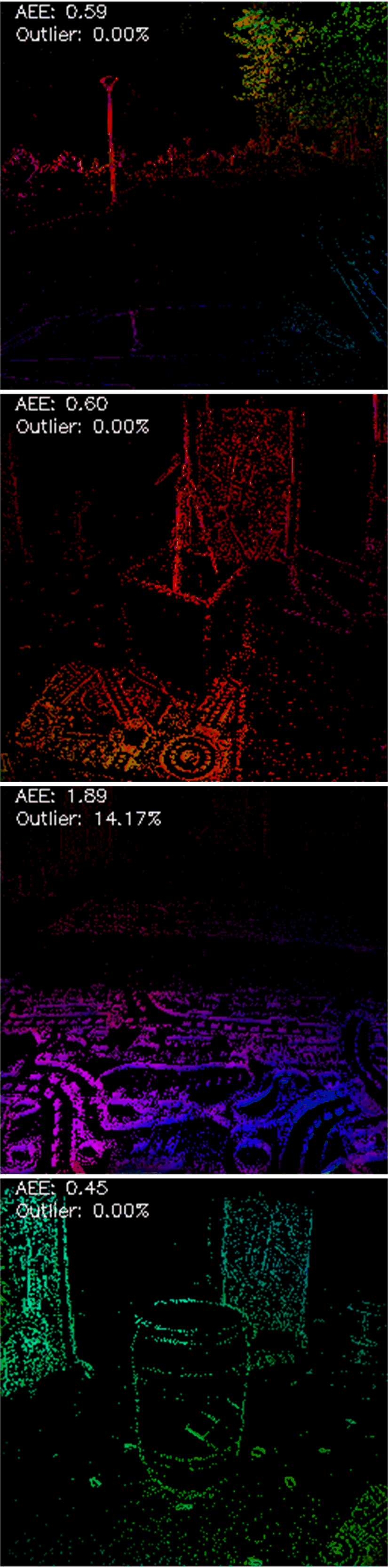}}
\hspace{-0.6em}
\subfigure[]{\label{fig:4d}\includegraphics[width=0.141\textwidth]{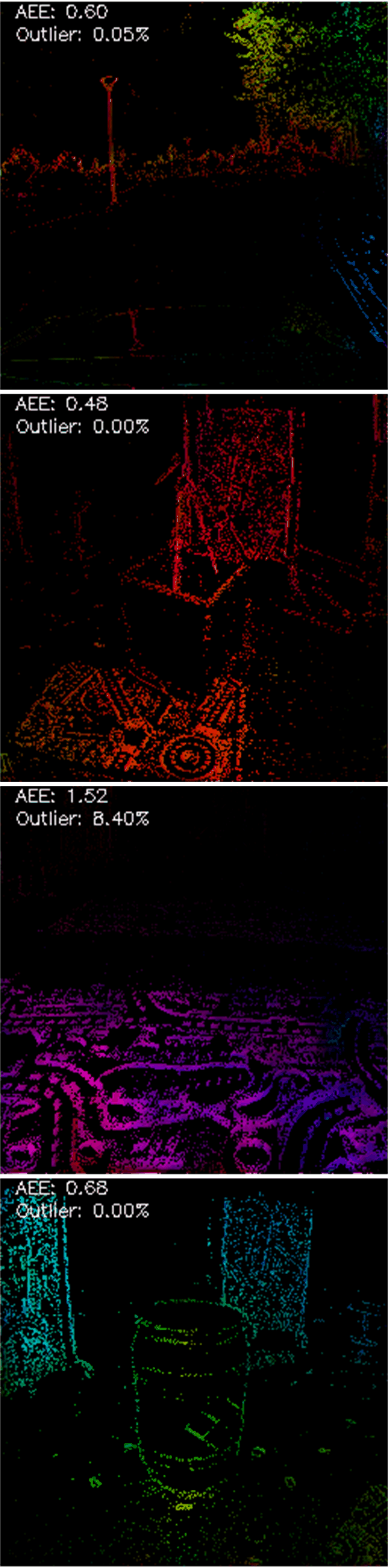}}
\hspace{-0.6em}
\subfigure[]{\label{fig:4e}\includegraphics[width=0.141\textwidth]{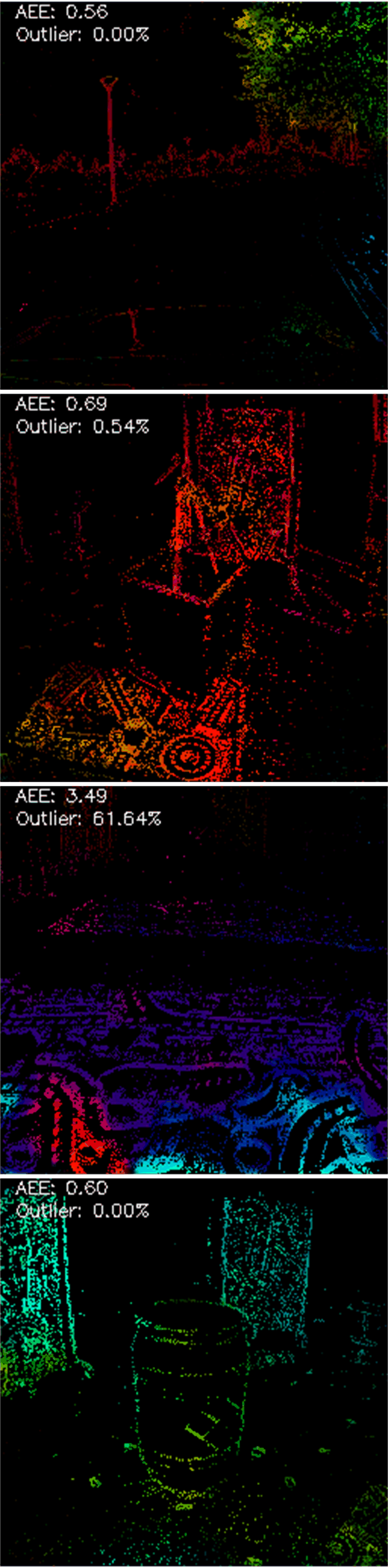}}
\hspace{-0.6em}
\subfigure[]{\label{fig:4f}\includegraphics[width=0.141\textwidth]{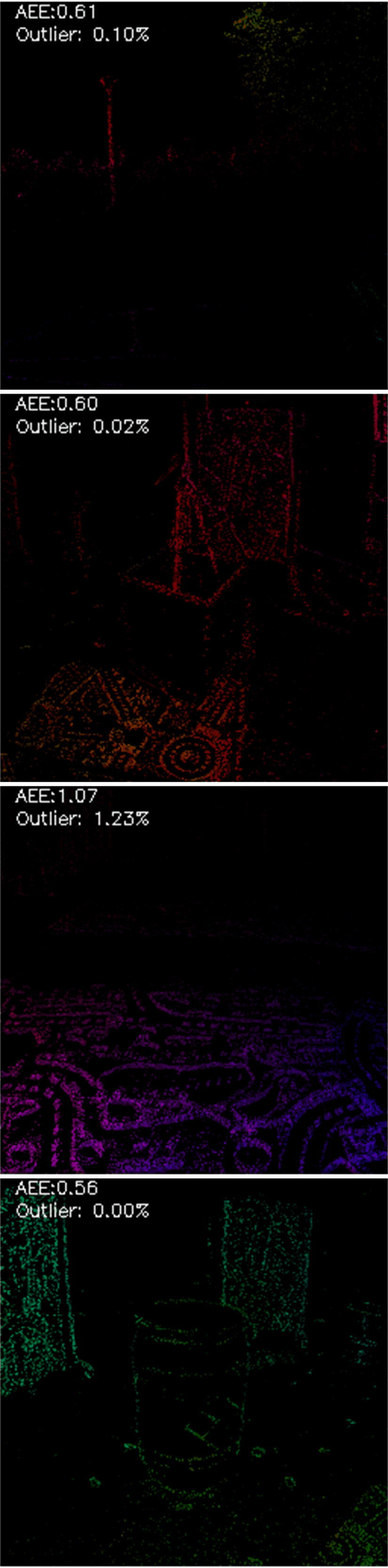}}
\hspace{-0.6em}
\subfigure[]{\label{fig:4g}\includegraphics[width=0.141\textwidth]{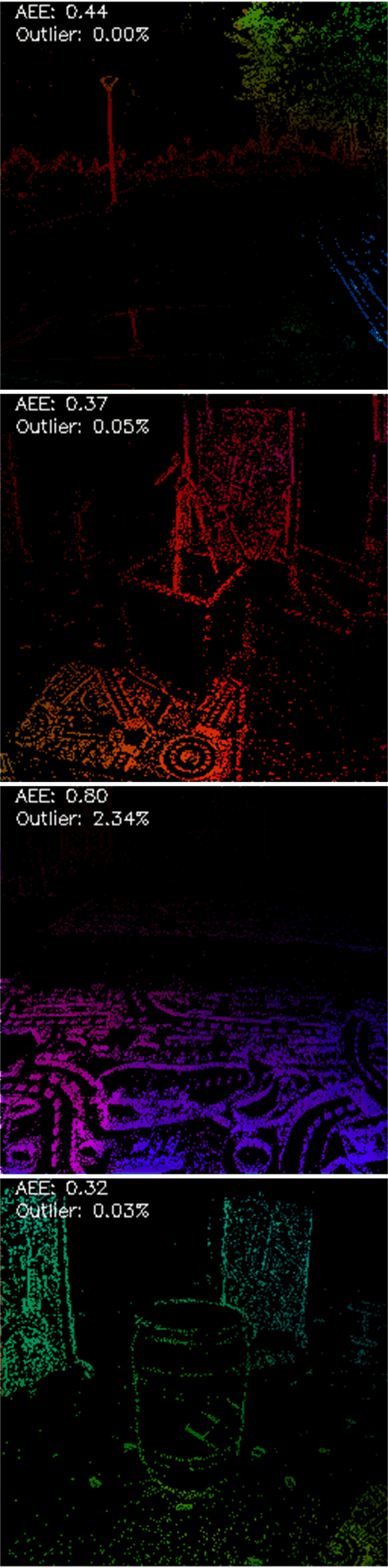}}
\vspace{-0.5em}
\caption{Qualitative evaluation of our best model compared to SOTA models on sequences from the MVSEC \cite{zhu2018multivehicle}. From top to bottom: outdoor\_day1, indoor\_flying1, indoor\_flying2, indoor\_flying3. (a) Events, (b) GT, (c) EV-FlowNet+ \cite{zhu2019unsupervised}, (d) ConvGRU-EV-FlowNet \cite{hagenaars2021self}, (e) XLIF-EV-FlowNet \cite{hagenaars2021self}, (f) STE-FlowNet \cite{ding2022spatio}, (g) EV-MGRFlowNet (Ours).}
\label{fig:5}
\vspace{-1em}
\end{figure*}

\subsection{Comparison against SOTA Methods}
\begin{figure*}[htbp]
\centering
\subfigure[]{\label{fig:5a}\includegraphics[width=\textwidth]{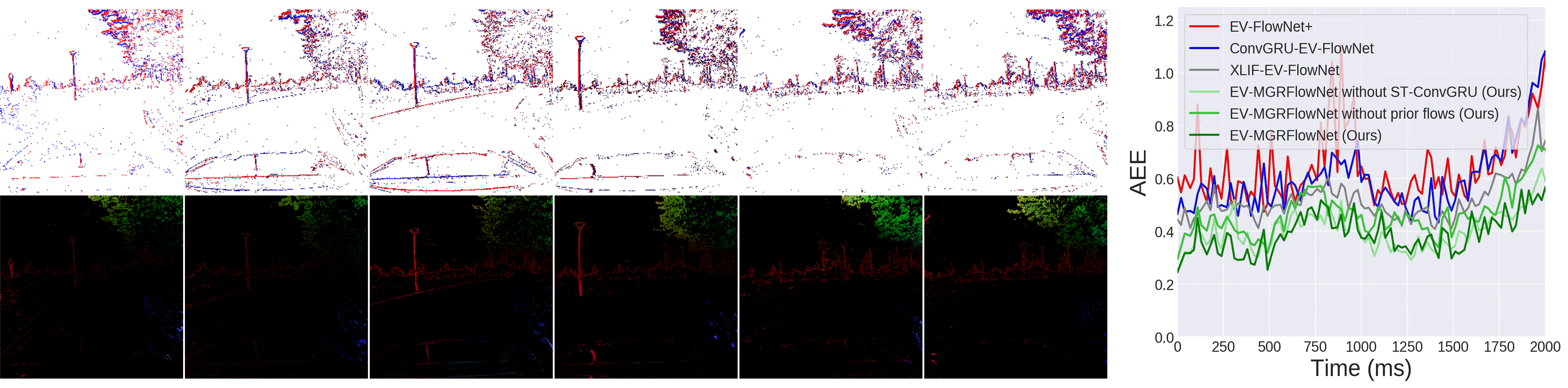}}
\subfigure[]{\label{fig:5b}\includegraphics[width=\textwidth]{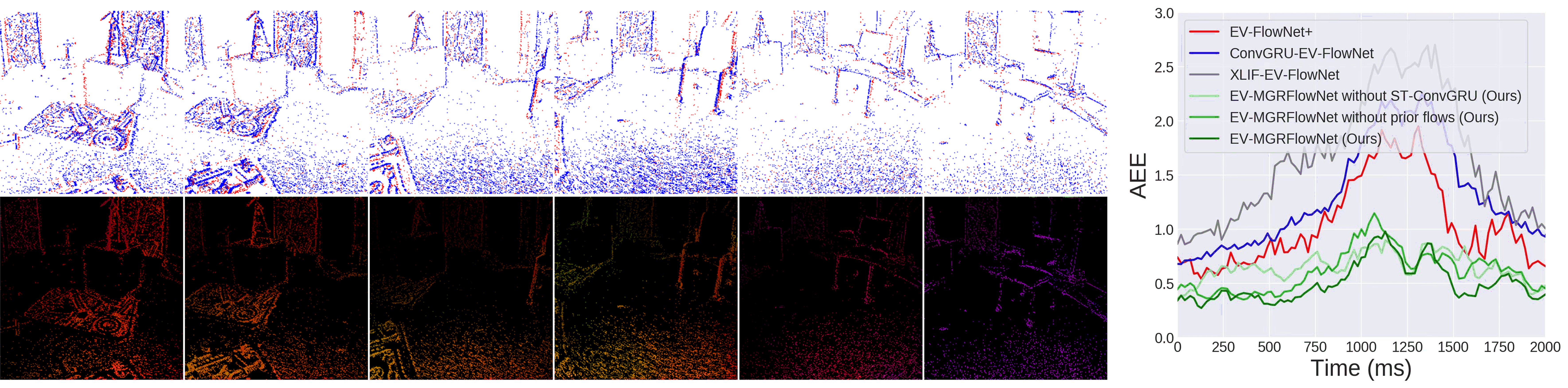}}
\vspace{-1em}
\caption{Typical examples of optical flow estimated by our EV-MGRFlowNet on some fragments of outdoor\_day1 and indoor\_flying1 from the MVSEC \cite{zhu2018multivehicle} dataset. We also show AEE curves between our EV-MGRFlowNet (including relevant ablation models) and several existing methods (EV-FlowNet+\cite{zhu2019unsupervised}, ConvGRU-EV-FlowNet\cite{hagenaars2021self}, XLIF-EV-FlowNet\cite{hagenaars2021self}). (a) The scene where a car is turning in outdoor\_day1 (from 4.4s to 6.4s). (b) The scene where a drone is making a sudden turn in indoor\_flying2 (from 40s to 42s).}
\label{fig:6}
\end{figure*}

We compare our EV-MGRFlowNet with existing event-based optical flow methods (EV-FlowNet \cite{zhu2018ev}, EV-FlowNet+ \cite{zhu2019unsupervised}, Spike-FlowNet \cite{lee2020spike}, XLIF-EV-FlowNet \cite{hagenaars2021self}, ConvGRU-EV-FlowNet \cite{hagenaars2021self}, ET-FlowNet \cite{tian2022event}) and the current SOTA method STE-FlowNet \cite{ding2022spatio}. The quantitative results in Table \ref{tab:1} demonstrate that our method outperforms the above methods in most cases. In the case of $\mathrm{d}t = 1$, compared to STE-FlowNet, our EV-MGRFlowNet reduces AEE by 33.33\%, 28.07\%, 11.39\% and 18.06\% on outdoor\_day1, indoor\_flying1, indoor\_flying2 and indoor\_flying3, respectively. As a whole, the AEE of our EV-MGRFlowNet is 22.71\% lower than that of STE-FlowNet on average. In the case of $\mathrm{d}t = 4$, The AEE and Outlier (\%) of EV-MGRFlowNet are 4.23\% and 2.32\% lower than those of STE-FlowNet on average, respectively. In summary, Our EV-MGRFlowNet achieves the lowest AEE on most sequences and the lower Outlier (\%) on multiple sequences. Meanwhile, the event count images, ground-truth flows and their corresponding estimated optical flows obtained from EV-FlowNet+ \cite{zhu2019unsupervised}, ConvGRU-EV-FlowNet \cite{hagenaars2021self}, XLIF-EV-FlowNet \cite{hagenaars2021self}, STE-FlowNet \cite{ding2022spatio} and our EV-MGRFlowNet on the aforementioned four sequences are visualized in Fig. \ref{fig:5}. In addition, the AEE curves in Fig. \ref{fig:6} also illustrate that our EV-MGRFlowNet achieves better results than other typical methods. These qualitative results also illustrate that our EV-MGRFlowNet can predict more accurate optical flow than other methods. Due to introducing prior hidden states and flows, our EV-MGRFlowNet can adequately capture spatio-temporal association between neighboring events. Meanwhile, our EV-MGRFlowNet uses HMC-Loss to strengthen geometric constraints for event alignment, thus unlocking the network’s potential through unsupervised learning. Therefore, as shown in Table \ref{tab:1} and Fig. \ref{fig:5}, our EV-MGRFlowNet can achieve SOTA performance with motion-guided recurrent networks using the hybrid motion-compensation loss. For a more detailed analysis of our network architecture and training paradigm, see Table \ref{tab:2}, Table \ref{tab:3} and Table \ref{tab:4}.

\begin{figure*}[htbp]
\centering
\subfigure[]{\label{fig:7a}\includegraphics[width=0.199\textwidth]{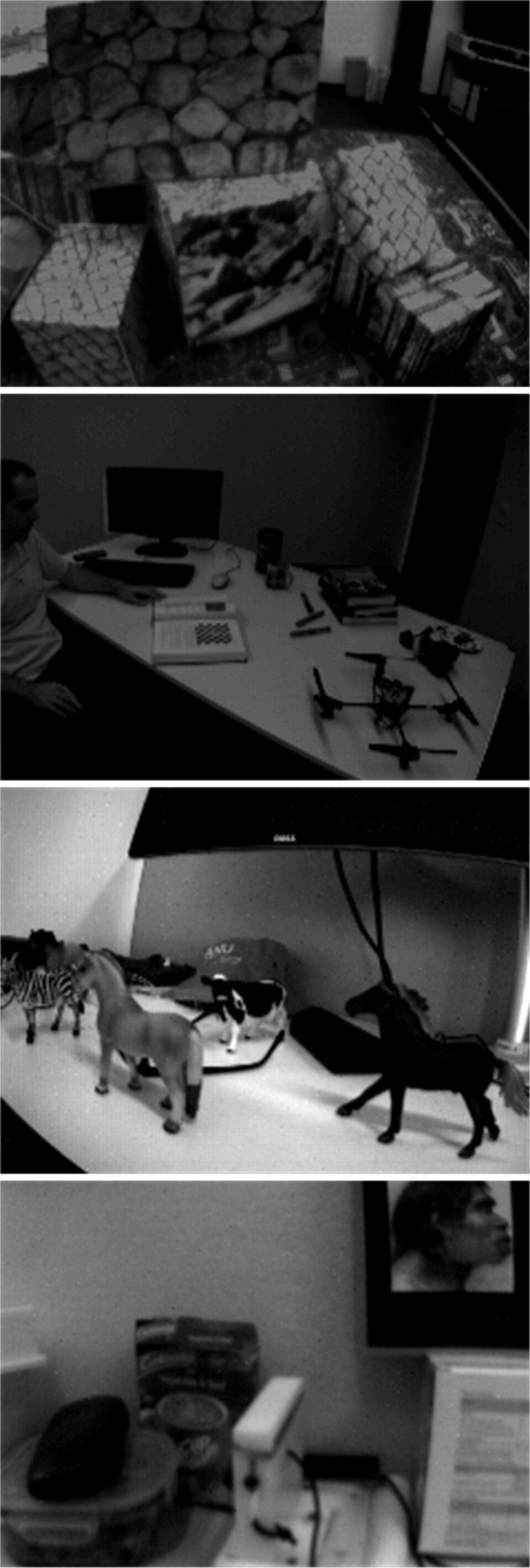}}
\hspace{-0.6em}
\subfigure[]{\label{fig:7b}\includegraphics[width=0.199\textwidth]{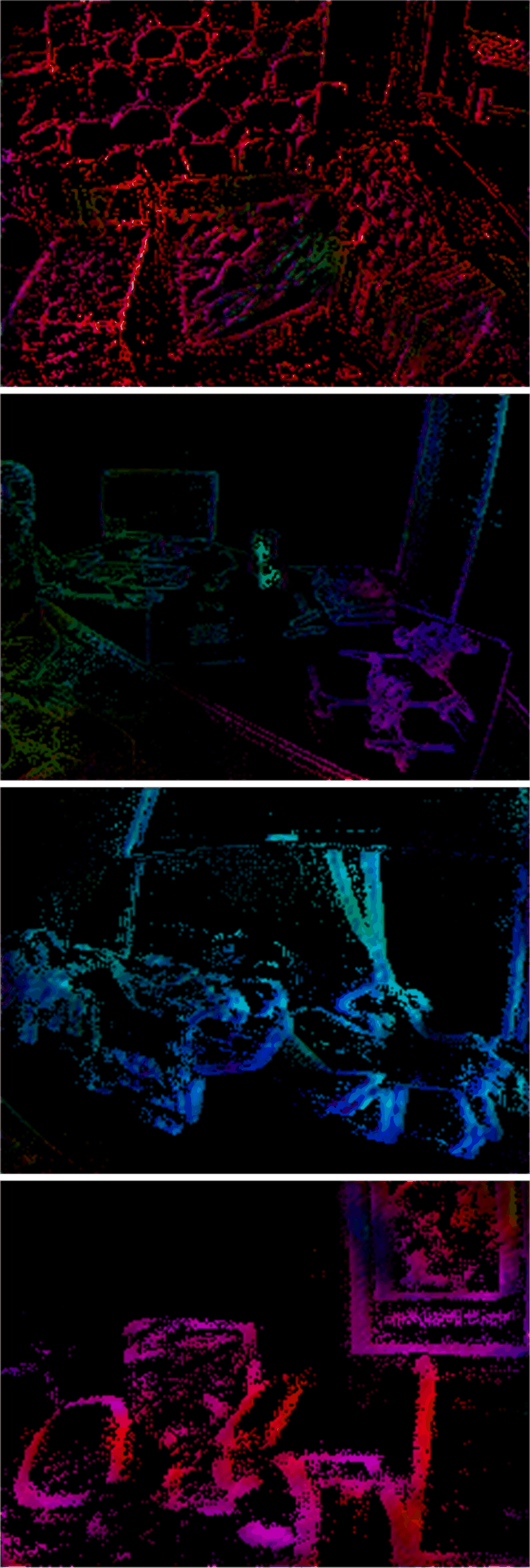}}
\hspace{-0.6em}
\subfigure[]{\label{fig:7c}\includegraphics[width=0.199\textwidth]{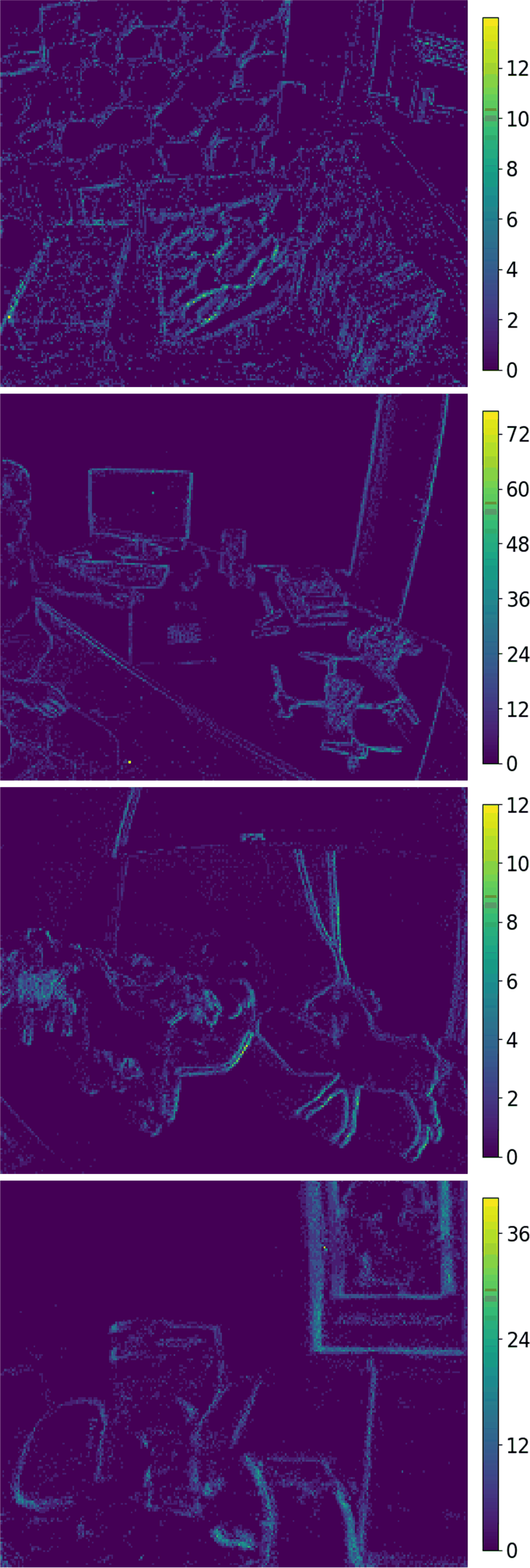}}
\hspace{-0.6em}
\subfigure[]{\label{fig:7d}\includegraphics[width=0.199\textwidth]{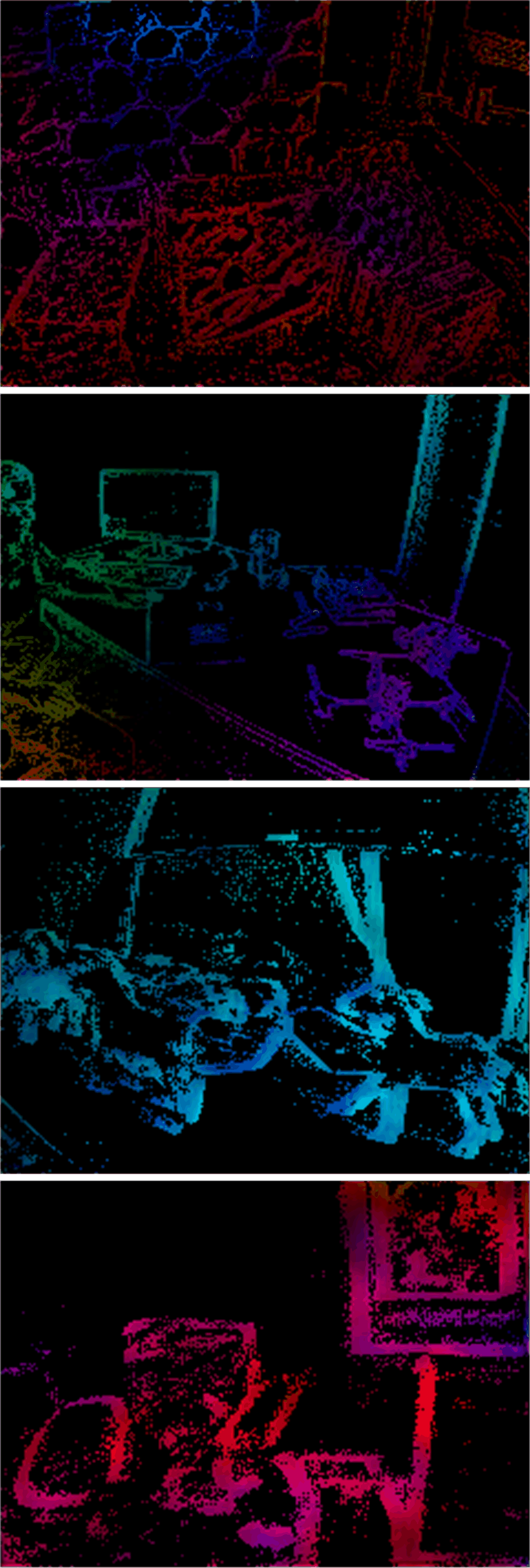}}
\hspace{-0.6em}
\subfigure[]{\label{fig:7e}\includegraphics[width=0.199\textwidth]{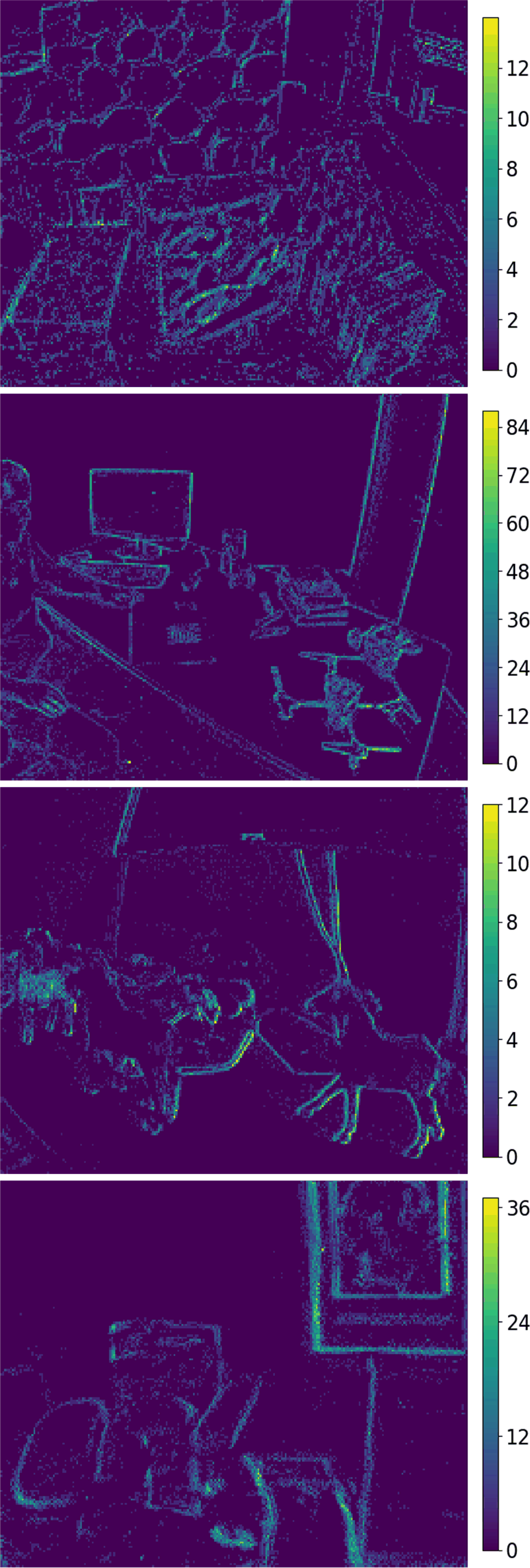}}
\vspace{-1em}
\caption{Comparison results of Optical Flow and IWE supervised by loss functions $L_{\text{AT}}$ and $L_{\text{HMC}}$ respectively based on ConvGRU-EV-FlowNet \cite{hagenaars2021self} on sequences from ECD\cite{mueggler2017event} and HQF\cite{stoffregen2020reducing}. From top to bottom: boxes\_6dof, dynamic\_6dof, desk\_fast, slow\_hand. (a) Intensity Frame. (b)-(c) Flow and IWE supervised by $L_{\text{AT}}$. (d)-(e) Flow and IWE supervised by $L_{\text{HMC}}$.}
\label{fig:7}
\end{figure*}

\begin{table*}[htbp]
\begin{center}
\caption{Ablation studies on FERE-Net and FGD-Net on the performance of EV-MGRFlowNet on MVSEC \cite{zhu2018multivehicle} with the best result \textbf{bolded}. FERE-Net: without/with ST-ConvGRU, FGD-Net: without/with prior flows.}
\newcommand{\tabincell}[2]{\begin{tabular}{@{}#1@{}}#2\end{tabular}}
\setlength{\tabcolsep}{0.005\linewidth}
\begin{tabular}{ccccccccccc p{1cm}}
\toprule
\multirow{3}{*}{\tabincell{c}{Ablation\\Studies}}
&\multirow{3}{*}{\tabincell{c}{Frame\\Interval}}
&\multirow{3}{*}{\tabincell{c}{Network Settings}}
&\multicolumn{8}{c}{MVSEC\cite{zhu2018multivehicle} Dataset}    \\
\cline{4-11} 
&
&
&\multicolumn{2}{c}{outdoor\_day1}
&\multicolumn{2}{c}{indoor\_flying1}
&\multicolumn{2}{c}{indoor\_flying2}
&\multicolumn{2}{c}{indoor\_flying3}   \\
\cline{4-11}
&   &   &AEE$\downarrow$    &Outlier (\%)$\downarrow$   &AEE$\downarrow$    &Outlier (\%)$\downarrow$   &AEE$\downarrow$    &Outlier (\%)$\downarrow$   &AEE$\downarrow$    &Outlier (\%)$\downarrow$   \\
\midrule
\multirow{4}{*}{FERE-Net}
&\multirow{2}{*}{$\mathrm{d}t=1$}   &without ST-ConvGRU &0.32   &0.14   &0.45   &0.34   &0.80   &3.33   &0.67   &2.24   \\
\cline{3-11} 
&   &with ST-ConvGRU    &\textbf{0.28}  &\textbf{0.02}  &\textbf{0.41}  &\textbf{0.17}  &\textbf{0.70}  &\textbf{2.35}  &\textbf{0.59}  &\textbf{1.29}  \\
\cline{2-11}
&\multirow{2}{*}{$\mathrm{d}t=4$}   &without ST-ConvGRU &1.19   &6.42   &1.67   &11.54  &2.73   &27.73  &2.24   &19.97  \\
\cline{3-11} 
&   &with ST-ConvGRU    &\textbf{1.10}  &\textbf{6.22}  &\textbf{1.50}  &\textbf{8.67}  &\textbf{2.39}  &\textbf{23.70} &\textbf{2.06}  &\textbf{18.00} \\
\midrule
\multirow{4}{*}{FGD-Net}
&\multirow{2}{*}{$\mathrm{d}t=1$}   &without prior flows    &0.30   &0.08   &0.41   &0.17   &0.78   &3.55   &0.63   &1.86   \\
\cline{3-11} 
&   &with prior flows   &\textbf{0.28}  &\textbf{0.02}  &\textbf{0.41}  &\textbf{0.17}  &\textbf{0.70}  &\textbf{2.35}  &\textbf{0.59}  &\textbf{1.29}  \\
\cline{2-11}
&\multirow{2}{*}{$\mathrm{d}t=4$}   &without prior flows  &1.20   &7.33   &1.53   &9.13   &2.67   &26.70  &2.13   &19.42  \\
\cline{3-11} 
&   &with prior flows    &\textbf{1.10}  &\textbf{6.22}  &\textbf{1.50}   &\textbf{8.67}  &\textbf{2.39}  &\textbf{23.70}  &\textbf{2.06}  &\textbf{18.00} \\
\bottomrule
\end{tabular}
\label{tab:2}
\vspace{-1.5em}
\end{center}
\end{table*}

\begin{table*}[htbp]
\begin{center}
\caption{Ablation studies on loss functions $L_{\text{AT}}$ \cite{hagenaars2021self} and $L_{\text{HMC}}$ respectively based on various networks on MVSEC\cite{zhu2018multivehicle}. We report AEE and Outlier (\%) for evaluation with the best results \textbf{bolded}.}
\newcommand{\tabincell}[2]{\begin{tabular}{@{}#1@{}}#2\end{tabular}}
\setlength{\tabcolsep}{0.005\linewidth}
\begin{tabular}{ccccccccccc p{1cm}}
\toprule
\multirow{3}{*}{\tabincell{c}{Frame\\Interval}}
&\multirow{3}{*}{Network Settings}
&\multirow{3}{*}{Loss}
&\multicolumn{8}{c}{MVSEC\cite{zhu2018multivehicle} Dataset}  \\
\cline{4-11}
&   &
&\multicolumn{2}{c}{outdoor\_day1}
&\multicolumn{2}{c}{indoor\_flying1}
&\multicolumn{2}{c}{indoor\_flying2}
&\multicolumn{2}{c}{indoor\_flying3}   \\
\cline{4-11}
&   &   
&AEE$\downarrow$    
&Outlier (\%)$\downarrow$   
&AEE$\downarrow$    
&Outlier (\%)$\downarrow$   
&AEE$\downarrow$    
&Outlier (\%)$\downarrow$   
&AEE$\downarrow$    
&Outlier (\%)$\downarrow$   \\
\midrule
\multirow{6}{*}{$\mathrm{d}t=1$}
&\multirow{2}{*}{ConvGRU-EV-FlowNet\cite{hagenaars2021self}} & $L_{\text{AT}}$   &0.47   &0.25   &0.60   &0.51   &1.17   &8.06   &0.93   &5.64   \\
\cline{3-11}
&   &$L_{\text{HMC}}$ &\textbf{0.32}  &\textbf{0.07}  &\textbf{0.49}  &\textbf{0.32}  &\textbf{0.85}  &\textbf{3.88}  &\textbf{0.69}  &\textbf{2.39}  \\ 
\cline{2-11}
&\multirow{2}{*}{XLIF-EV-FlowNet\cite{hagenaars2021self}}   &$L_{\text{AT}}$ &0.45   &0.16   &0.73   &0.92   &1.45   &12.18   &1.17   &8.35   \\
\cline{3-11}
&   &$L_{\text{HMC}}$ &\textbf{0.40}  &\textbf{0.12}  &\textbf{0.65}  &\textbf{0.33}  &\textbf{1.39}  &\textbf{10.37}  &\textbf{1.10}  &\textbf{6.92}  \\
\cline{2-11}
&\multirow{2}{*}{EV-MGRFlowNet (Ours)}   &$L_{\text{AT}}$   &0.37   &0.07   &0.52   &0.33   &0.96   &5.67   &0.79   &3.87   \\
\cline{3-11}
&   &$L_{\text{HMC}}$ &\textbf{0.28}  &\textbf{0.02}  &\textbf{0.41}  &\textbf{0.17}  &\textbf{0.70}  &\textbf{2.35}  &\textbf{0.59}  &\textbf{1.29}  \\
\midrule
\multirow{6}{*}{$\mathrm{d}t=4$}
&\multirow{2}{*}{ConvGRU-EV-FlowNet\cite{hagenaars2021self}} &$L_{\text{AT}}$ &1.69   &12.50  &2.16   &21.51  &3.90   &40.72  &3.00   &29.60  \\
\cline{3-11}
&   &$L_{\text{HMC}}$ &\textbf{1.26}  &\textbf{6.16}  &\textbf{1.77}  &\textbf{13.52} &\textbf{2.92}  &\textbf{30.91} &\textbf{2.32}  &\textbf{22.67} \\
\cline{2-11} 
&\multirow{2}{*}{XLIF-EV-FlowNet\cite{hagenaars2021self}}   &$L_{\text{AT}}$    &1.67   &12.69   &2.72   &31.69  &4.93   &\textbf{51.36}  &3.91   &42.52  \\ 
\cline{3-11} 
&   &$L_{\text{HMC}}$ &\textbf{1.55}  &\textbf{8.25}  &\textbf{2.44}  &\textbf{27.13}  &\textbf{4.78}  &52.35 &\textbf{3.71}  &\textbf{41.90} \\ 
\cline{2-11} 
&\multirow{2}{*}{EV-MGRFlowNet (Ours)}   &$L_{\text{AT}}$  &1.41   &9.09   &1.90   &16.70  &3.13   &33.05  &2.58   &25.67  \\ 
\cline{3-11} 
&   &$L_{\text{HMC}}$ &\textbf{1.10}  &\textbf{6.22}  &\textbf{1.50}  &\textbf{8.67}  &\textbf{2.39}  &\textbf{23.70} &\textbf{2.06}  &\textbf{18.00} \\ 
\bottomrule
\end{tabular}
\label{tab:3}
\vspace{-1.0em}
\end{center}
\end{table*}

\begin{table*}[htbp]
\begin{center}
\caption{Ablation studies on loss functions $L_{\text{AT}}$ \cite{hagenaars2021self} and $L_{\text{HMC}}$ respectively based on ConvGRU-EV-FlowNet \cite{hagenaars2021self} on ECD\cite{mueggler2017event} and HQF\cite{stoffregen2020reducing}. We report FWL and RSAT for evaluation with the best results \textbf{bolded}.}
\newcommand{\tabincell}[2]{\begin{tabular}{@{}#1@{}}#2\end{tabular}}
\setlength{\tabcolsep}{0.005\linewidth}
\begin{tabular}{ccccccccccccccccc p{1cm}}
\toprule
\multirow{3}{*}{Loss}
&\multicolumn{8}{c}{ECD\cite{mueggler2017event} Dataset}
&\multicolumn{8}{c}{HQF\cite{stoffregen2020reducing} Dataset}  \\
\cline{2-17}
&\multicolumn{2}{c}{dynamic\_6dof}
&\multicolumn{2}{c}{boxes\_6dof}
&\multicolumn{2}{c}{poster\_6dof}
&\multicolumn{2}{c}{calibration}
&\multicolumn{2}{c}{desk\_fast}
&\multicolumn{2}{c}{slow\_hand}
&\multicolumn{2}{c}{poster\_pillar\_2}
&\multicolumn{2}{c}{still\_life}   \\
\cline{2-17}
&FWL$\uparrow$  &RAST$\downarrow$   &FWL$\uparrow$  &RAST$\downarrow$   &FWL$\uparrow$  &RAST$\downarrow$   &FWL$\uparrow$  &\multicolumn{1}{c}{RAST$\downarrow$}   &FWL$\uparrow$  &RAST$\downarrow$   &FWL$\uparrow$  &RAST$\downarrow$   &FWL$\uparrow$  &RAST$\downarrow$   &FWL$\uparrow$  &RAST$\downarrow$   \\
\midrule
$L_\text{AT}$   &1.39   &0.90   &1.58   &0.93   &1.55   &0.93   &1.09   &0.96   &1.37   &0.87 &1.49   &0.95   &1.06   &0.95   &1.50   &0.93  \\
$L_\text{HMC}$   &\textbf{1.46}  &\textbf{0.88}  &\textbf{1.64}  &\textbf{0.92}  &\textbf{1.62}  &\textbf{0.91}  &\textbf{1.13}  &\textbf{0.94}  &\textbf{1.48}  &\textbf{0.85}  &\textbf{1.57}  &\textbf{0.93}  &\textbf{1.10}  &\textbf{0.92}  &\textbf{1.60}  &\textbf{0.91}   \\
\bottomrule
\end{tabular}
\label{tab:4}
\vspace{-1.0em}
\end{center}
\end{table*}

\subsection{Ablation Studies}
In this section, we explore the impact of FERE-Net, FGD-Net and HMC-Loss (denoted as$L_{\text{HMC}}$) on the performance of EV-MGRFlowNet on the MVSEC dataset.
\subsubsection{Impact of FERE-Net}
Given that ST-ConvGRU is the key component of our proposed FERE-Net, we design a similar encoder network where ST-ConvGRU is replaced with standard ConvGRU for ablation study. The comparison results in Table \ref{tab:2} show that incorporating ST-ConvGRU in the FERE-Net reduces AEE by 11.46\% and 9.56\% in the case of $\mathrm{d}t=1$ and $\mathrm{d}t=4$, respectively. The AEE curves in Fig. \ref{fig:6} also illustrate that our EV-MGRFlowNet achieves improved accuracy in predicting optical flow by incorporating ST-ConvGRU. These results demonstrate that introducing high-level hidden states into the low-level encoding layer can provide significant motion cues, and our proposed ST-ConvGRU can help FERE-Net fully utilize multi-level hidden states. 

\subsubsection{Impact of FGD-Net}
We design a similar decoder network without introducing prior flows for the ablation study. The comparison results in Table \ref{tab:2} show that the performance of our network can be further enhanced with FGD-Net. In more detail, introducing prior flows in the FGD-Net reduces AEE by 5.82\% and 6.02\% in the case of $\mathrm{d}t=1$ and $\mathrm{d}t=4$, respectively. The AEE curves in Fig. \ref{fig:6} also illustrate that our EV-MGRFlowNet yields more precise optical flow by incorporating prior flows. These results demonstrate that prior flows are helpful for the current optical flow estimation, and our proposed FGD-Net can effectively refine the optical flows. 

\subsubsection{Impact of HMC-Loss}
Here, to comprehensively validate the effectiveness of HMC-Loss, we use our $L_{\text{HMC}}$ and the widely used loss $L_{\text{AT}}$ \cite{zhu2019unsupervised,hagenaars2021self} respectively to train various networks on multiple datasets for the ablation study. Firstly, as shown in Table \ref{tab:3}, we evaluated various network architectures on MVSEC \cite{zhu2018multivehicle}, including ConvGRU-EV-FlowNet\cite{hagenaars2021self}, XLIF-EV-FlowNet\cite{hagenaars2021self} and our EV-MGRFlowNet. Compared to $L_{\text{AT}}$, training ConvGRU-EV-FlowNet, XLIF-EV-FlowNet and EV-MGRFlowNet using $L_{\text{HMC}}$ reduces AEE by 25.85\% and 22.83\% for ConvGRU-EV-FlowNet, 8.05\% and 6.41\% for XLIF-EV-FlowNet and 24.47\% and 21.71\% for EV-MGRFlowNet in the case of $\mathrm{d}t=1$ and $\mathrm{d}t=4$, respectively. The above results show that our loss function does not depend on the specific network architectures. Moreover, as shown in Table \ref{tab:4}, we evaluate ConvGRU-EV-FlowNet\cite{hagenaars2021self} on the ECD \cite{mueggler2017event} and HQF \cite{stoffregen2020reducing} datasets. To make a fair comparison with the experimental data in \cite{hagenaars2021self}, we choose their ConvGRU-EV-FlowNet instead of our EV-MGRFlowNet as the network architecture. Note that we use the Flow Warp Loss (FWL)\cite{stoffregen2020reducing} and the Ratio of the Squared Average Timestamps (RSAT)\cite{hagenaars2021self} for evaluation. The quantitative results in Table \ref{tab:4} show that training ConvGRU-EV-FlowNet using $L_{\text{HMC}}$ achieves better results than using $L_{\text{AT}}$ in both FWL and RSAT metrics. These two metrics can reflect the quality of optical flow by measuring the alignment of events warped by optical flow. The optical flow and IWE in Fig. \ref{fig:7} also illustrate that our proposed $L_{\text{HMC}}$ can bring better event alignment and gain sharper count IWEs, thus accurately estimating optical flow. For more details, please refer to previous works \cite{stoffregen2020reducing,hagenaars2021self} about these metrics. These results show that our loss function does not depend on the specific evaluation datasets. In summary, the comparison results ($L_{\text{AT}}$ v.s. $L_{\text{HMC}}$) in both Table \ref{tab:3} and Table \ref{tab:4} show that the trained models using $L_{\text{HMC}}$ substantially outperform the corresponding ones using $L_{\text{AT}}$. These results demonstrate that our proposed $L_{\text{HMC}}$ strengthens geometric constraints for accurate event alignment, thus unlocking the potential of networks through unsupervised learning.

\section{Conclusions}
\label{sec:conclusions}
In this paper, we analyze the limitations of previous event-based optical flow estimation methods from the perspective of network architecture and training paradigm. Based on that, we propose a method for unsupervised event-based optical flow estimation with motion-guided networks using a novel hybrid motion-compensation loss (EV-MGRFlowNet). The two key ideas of our EV-MGRFlowNet are as follows: First, we propose a feature-enhanced recurrent encoder network (FERE-Net) that incorporates a novel memory unit, ST-ConvGRU, to fully utilize multi-level hidden states. Next, we propose a flow-guided decoder network (FGD-Net) to fuse the previously estimated optical flow with the current one to achieve better results. Then, we propose a novel hybrid motion-compensation loss (HMC-Loss), which strengthens geometric constraints for better event alignment. Hence, our proposed loss function unlocks the network's potential through unsupervised learning, thus substantially improving the accuracy of optical flow estimation. The experimental results demonstrate that our method outperforms all existing SOTA methods in indoor/outdoor scenes. In the future, we will explore the realization of a fusion network that combines events and frames for higher performance \cite{hou2023fe} and a deep spiking optical flow network architecture for highly energy-efficient inference \cite{jiang2022neuro}.
 
\bibliographystyle{IEEEtran}
\bibliography{mybibfile}

\begin{IEEEbiography}[{\includegraphics[width=1in,height=1.25in,clip,keepaspectratio]{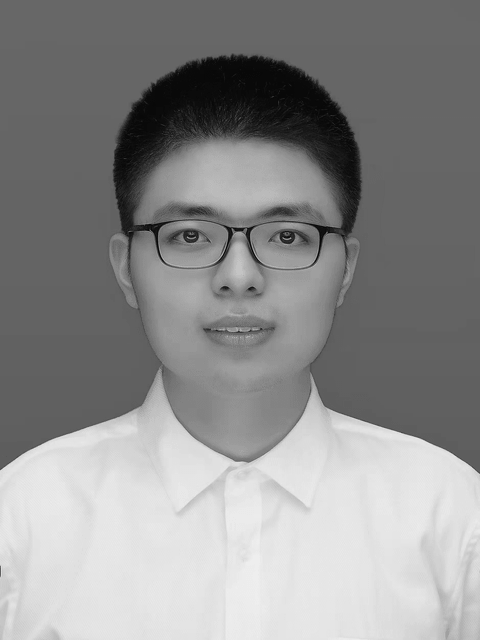}}]{Hao Zhuang}
received the B.S. degree in mechanical and electronic engineering with Chongqing University of Posts and Telecommunications, Chongqing, China, in 2021. He is currently pursuing the M.S. degree in control engineering with Northeastern University, Shenyang, China. His research interests include event-based vision, optical flow estimation and deep learning.\end{IEEEbiography}
\vspace{-3em}
\begin{IEEEbiography}[{\includegraphics[width=1in,height=1.25in,clip,keepaspectratio]{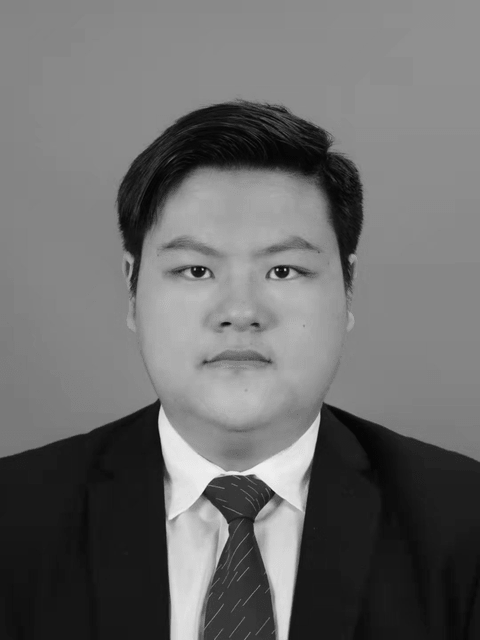}}]{Xinjie Huang}
received the B.S. degree in automation with Zhejiang University of Technology, Hangzhou, China, in 2021. He is currently pursuing the M.S. degree in robot science and engineering with Northeastern University, Shenyang, China. His research interests include event-based vision, depth estimation and deep learning.\end{IEEEbiography}
\vspace{-3em}
\begin{IEEEbiography}[{\includegraphics[width=1in,height=1.25in,clip,keepaspectratio]{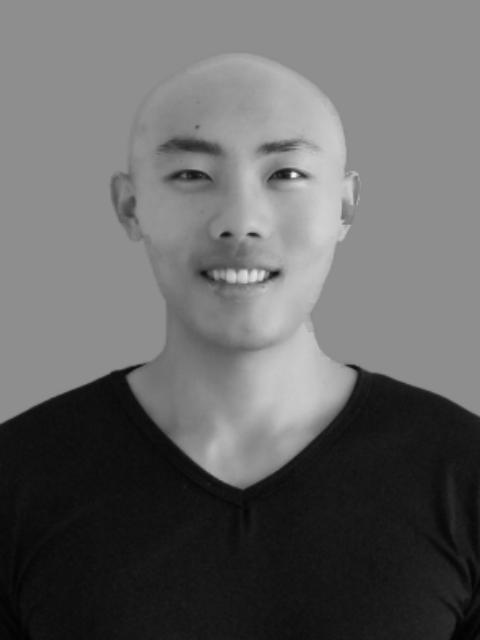}}]{Kuanxu Hou}
received the B.S. degree in robot engineering from Northeastern University, Shenyang, China, in 2020. He is currently pursuing the M.S. degree in robot science and engineering with Northeastern University, Shenyang, China. His research interests include event-based vision, visual place recognition and deep learning.\end{IEEEbiography}
\vspace{-3em}
\begin{IEEEbiography}[{\includegraphics[width=1in,height=1.25in,clip,keepaspectratio]{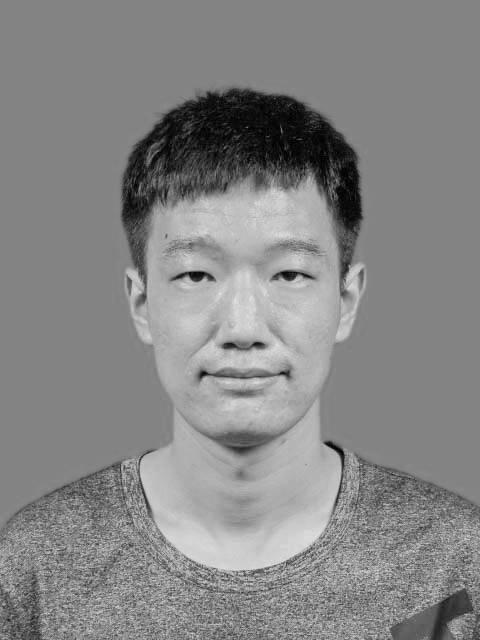}}]{Delei Kong}
received the B.S. degree in automation from Henan Polytechnic University, Zhengzhou, China, in 2018 and the M.S. degree in control engineering from Northeastern University, Shenyang, China, in 2021. From 2020 to 2021, he worked as  an Algorithm Intern at SynSense Tech. Co. Ltd. From 2021 to 2022, he worked as an R\&D Engineer (advanced vision) at Machine Intelligence Laboratory, China Nanhu Academy of Electronics and Information Technology (CNAEIT). Since 2021, he also has been a Research Assistant at Northeastern University. His research interests include event-based vision, robot visual navigation and neuromorphic computing.\end{IEEEbiography}
\vspace{-3em}
\begin{IEEEbiography}[{\includegraphics[width=1in,height=1.25in,clip,keepaspectratio]{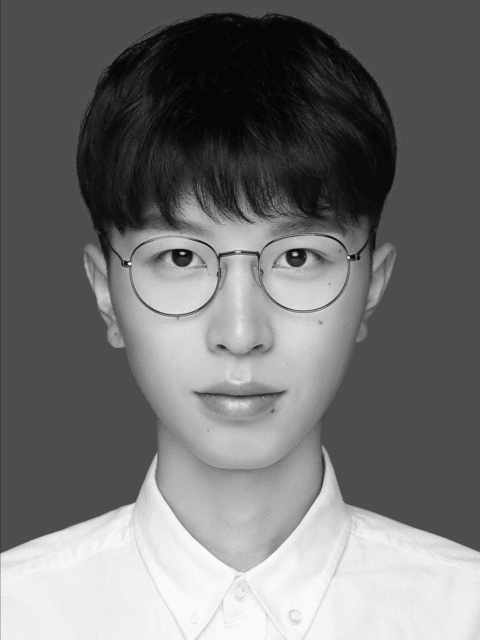}}]{Chenming Hu}
received the B.S. degree in robot engineering from Nanjing University of Information Science and Technology, Nanjing, China, in 2022. He is currently pursuing the M.S. degree in robot science and engineering with Northeastern University, Shenyang, China. His research interests include event-based vision, deep learning and neuromorphic computing.\end{IEEEbiography}
\vspace{-3em}
\begin{IEEEbiography}[{\includegraphics[width=1in,height=1.25in,clip,keepaspectratio]{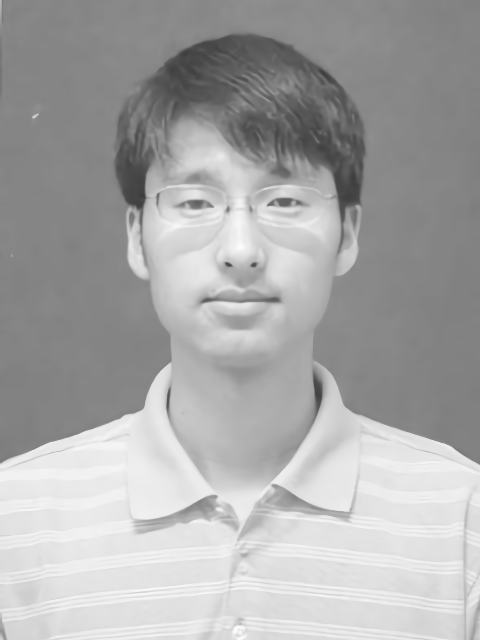}}]{Zheng Fang}
(Member, IEEE) received the B.S. degree in automation and the Ph.D. degree in pattern recognition and intelligent systems from Northeastern University, Shenyang, China, in 2002 and 2006, respectively. He was a Post-Doctoral Research Fellow at the Robotics Institute of Carnegie Mellon University (CMU), Pittsburgh, PA, USA, from 2013 to 2015. He is currently a Professor with the Faculty of Robot Science and Engineering, Northeastern University. He has published over 70 papers in well-known journals or conferences in robotics and computer vision, including Journal of Field Robotics (JFR), IEEE Transactions on Pattern Analysis and Machine Intelligence (TPAMI), IEEE Transactions on Multimedia (TMM), IEEE Robotics and Automation Letters (RA-L), IEEE International Conference on Robotics and Automation (ICRA), IEEE/RSJ International Conference on Intelligent Robots (IROS), and so on. His research interests include visual/laser simultaneous localization and mapping (SLAM), and perception and autonomous navigation of various mobile robots.\end{IEEEbiography}
\end{document}